\documentclass[final]{cvpr}

\usepackage{times}
\usepackage{epsfig}
\usepackage{graphicx}
\usepackage{xcolor}
\usepackage[font=normal]{subfig}
\usepackage{tabularx}
\usepackage{booktabs}
\usepackage{amsmath}
\usepackage{amssymb}
\usepackage[hang,flushmargin]{footmisc}

\usepackage[pagebackref=true,breaklinks=true,colorlinks,bookmarks=false,citecolor=blue]{hyperref}

\def\cvprPaperID{53} 
\def\confYear{CVPR 2021}

\newcommand{\tabcspace}{\vspace{-2.5mm}}
\newcommand{\tabspace}{\vspace{-4.5mm}}
\newcommand{\figcspace}{\vspace{-2.5mm}}
\newcommand{\figspace}{\vspace{-4.5mm}}

\newcommand{\Paragraph}[1]{\noindent\textbf{#1}}
\newcommand{\code}[1]{{\fontfamily{qcr}\selectfont #1}}

\begin{document}

\title{NTIRE 2021 Challenge on Video Super-Resolution}

\author{
    Sanghyun Son$^\dagger$ \and Suyoung Lee$^\dagger$ \and Seungjun Nah$^\dagger$ \and Radu Timofte$^\dagger$ \and Kyoung Mu Lee$^\dagger$    
    
    \and Kelvin C.K. Chan \and Shangchen Zhou \and Xiangyu Xu \and Chen Change Loy          
    \and Boyuan Jiang \and Chuming Lin \and Yuchun Dong \and Donghao Luo \and Wenqing Chu   
    \and Xiaozhong Ji \and Siqian Yang                                                      
    \and Ying Tai \and Chengjie Wang \and Jilin Li \and Feiyue Huang                        
    \and Chengpeng Chen \and Xiaojie Chu \and Jie Zhang \and Xin Lu \and Liangyu Chen       
    \and Jing Lin \and Guodong Du \and Jia Hao \and Xueyi Zou                               
    \and Qi Zhang \and Lielin Jiang \and Xin Li \and He Zheng \and Fanglong Liu 
    \and Dongliang He \and Fu Li \and Qingqing Dang                                         
    \and Peng Yi \and Zhongyuan Wang \and Kui Jiang \and Junjun Jiang \and Jiayi Ma         
    \and Yuxiang Chen \and Yutong Wang                                                      
    \and Ting Liu \and Qichao Sun \and Huanwei Liang                                        
    \and Yiming Li \and Zekun Li \and Zhubo Ruan \and Fanjie Shang \and Chen Guo \and Haining Li
    \and Renjun Luo \and Longjie Shen                                                       
    \and Kassiani Zafirouli \and Konstantinos Karageorgos \and Konstantinos Konstantoudakis 
    \and Anastasios Dimou \and Petros Daras                                                 
    \and Xiaowei Song \and Xu Zhuo \and Hanxi Liu                                           
    
    \and Mengxi Guo \and Junlin Li                                                          
    \and Yu Li \and Ye Zhu                                                                  
    \and Qing Wang \and Shijie Zhao \and Xiaopeng Sun \and Gen Zhan 
    \and Tangxin Xie \and Yu Jia                                                            
    \and Yunhua Lu \and Wenhao Zhang \and Mengdi Sun \and Pablo Navarrete Michelini         
    \and Xueheng Zhang \and Hao Jiang \and Zhiyu Chen \and Li Chen                          
    \and Zhiwei Xiong \and Zeyu Xiao \and Ruikang Xu \and Zhen Cheng 
    \and Xueyang Fu \and Fenglong Song                                                      
    \and Zhipeng Luo \and Yuehan Yao                                                        
    \and Saikat Dutta \and Nisarg A. Shah \and Sourya Dipta Das                             
    \and Peng Zhao \and Yukai Shi \and Hongying Liu \and Fanhua Shang \and Yuanyuan Liu 
    \and Fei Chen \and Fangxu Yu \and Ruisheng Gao \and Yixin Bai                           
    \and Jeonghwan Heo                                                                      
    \and Shijie Yue \and Chenghua Li \and Jinjing Li \and Qian Zheng 
    \and Ruipeng Gang \and Ruixia Song                                                      
    \and Seungwoo Wee \and Jechang Jeong                                                    
    \and Chen Li \and Geyingjie Wen \and Xinning Chai \and Li Song                          
}

\maketitle

\begin{abstract}
    Super-Resolution~(SR) is a fundamental computer vision task that aims to obtain a high-resolution clean image from the given low-resolution counterpart.
    This paper reviews the NTIRE 2021 Challenge on Video Super-Resolution.
    We present evaluation results from two competition tracks as well as the proposed solutions.
    Track 1 aims to develop conventional video SR methods focusing on the restoration quality.
    Track 2 assumes a more challenging environment with lower frame rates, casting spatio-temporal SR problem.
    In each competition, 247 and 223 participants have registered, respectively.
    During the final testing phase, 14 teams competed in each track to achieve state-of-the-art performance on video SR tasks.
\end{abstract}

{\let\thefootnote\relax\footnotetext{
\noindent
$^\dagger$ S. Son (thstkdgus35@snu.ac.kr, Seoul National University), S. Lee, S. Nah, R. Timofte, K. M. Lee are the NTIRE 2021 challenge organizers, while the other authors participated in the challenge. 
\\Appendix~\ref{sec:appendix} contains the authors' teams and affiliations.
\\Website: \url{https://data.vision.ee.ethz.ch/cvl/ntire21/}}}

\section{Introduction}
\label{sec:intro}

With increasing demands for high-quality videos such as OTT service, live streaming, and personal media, improving the quality of videos is considered an important computer vision problem these days.
Among them, video super-resolution (VSR) aims to reconstruct a high-resolution (HR) sequence from the given low-resolution (LR) input frames.
Different from single image super-resolution (SISR), where input and output images are constrained to a single timestep, VSR can utilize temporal dynamics to deliver more accurate reconstruction.
Recent methods adopt explicit flow estimation~\cite{Caballero_2017_CVPR, Tao_2017_ICCV, xue2019video}, implicit alignment~\cite{jo2018deep, wang2019edvr} or attention~\cite{isobe2020video_cvpr} to effectively aggregate temporal information distributed over time.

In NTIRE 2021 Challenge on Video Super-Resolution, participants are required to develop state-of-the-art VSR methods on two distinctive tracks.
The goal of Track 1 is to reconstruct 30 HR sequences from given $\times 4$ downsampled videos from the REDS~\cite{Nah_2019_CVPR_Workshops_REDS} dataset, similar to the conventional VSR algorithms.
In Track 2, the input LR video has a lower frame rate, i.e., 12fps, than Track 1, which deals with 24fps sequences.
The participants are asked to interpolate the given frames across spatial and time domains jointly, which is called spatio-temporal super-resolution (STSR).

This challenge is one of the NTIRE 2021 associated challenges: nonhomogeneous dehazing~\cite{ancuti2021ntire}, defocus deblurring using dual-pixel~\cite{abuolaim2021ntire}, depth guided image relighting~\cite{elhelou2021ntire}, image deblurring~\cite{nah2021ntire}, multi-modal aerial view imagery classification~\cite{liu2021ntire}, learning the super-resolution space~\cite{lugmayr2021ntire}, quality enhancement of heavily compressed videos~\cite{yang2021ntire}, video super-resolution, perceptual image quality assessment~\cite{gu2021ntire}, burst super-resolution~\cite{bhat2021ntire}, high dynamic range~\cite{perez2021ntire}.

\section{Related Works}
\label{sec:related_works}
\Paragraph{Video super-resolution.}
Different from the SISR model, VSR utilizes neighboring frames in a video to reconstruct the high-resolution sequence.
Since images from different timesteps are not aligned in spatial domain, one of the primary interests of various VSR models is to predict an accurate alignment between adjacent frames.
Early approaches~\cite{Caballero_2017_CVPR, Tao_2017_ICCV} explicitly deal with the issue by flow estimation models in their framework.
In TOFlow~\cite{xue2019video}, a concept of task-oriented flow is proposed which is lighter and designed to handle various video processing problems.
However, such explicit flow models have several limitations as LR images may not contain enough information to predict the accurate flow.
Therefore, DUF~\cite{jo2018deep} introduce dynamic upsampling module without including explicit motion compensation in the VSR framework.

As the winner of NTIRE 2019 Video Super-Resolution and Deblurring Challenges~\cite{Nah_2019_CVPR_Workshops_SR, Nah_2019_CVPR_Workshops_Deblur, Nah_2019_CVPR_Workshops_REDS}, EDVR~\cite{wang2019edvr} combines several novel components such as PCD alignment and TSA fusion to achieve state-of-the-art reconstruction quality.
Recent methods adopt various techniques such as recurrent structure-detail network~\cite{isobe2020video}, multi-correspondence aggregation~\cite{li2020mucan}, temporal group attention~\cite{isobe2020video_cvpr}, or temporally deformable alignement~\cite{tian2018tdan}.
In Track 1 of NTIRE 2021 Video Super-Resolution Challenge, we promote participants to construct effective and novel frameworks for VSR problem.

\Paragraph{Video frame interpolation.}
Earlier methods are constructed on phase-based methods~\cite{Meyer_2015_CVPR, Meyer_2018_CVPR}, where the temporal change of neighboring frames is represented as shifts of phase.
Recently, most of the frame interpolation methods consider motion dynamics with neural networks rather than directly synthesize intermediate frames~\cite{long2016learning,choi2020channel}.
Among them, flow-based approaches are one of the most popular solutions.
DVF~\cite{Liu_2017_ICCV} and SuperSloMo~\cite{Jiang_2018_CVPR} adopt piece-wise linear models which estimate optical flow between two input frames and warp them to the target intermediate time.
The accuracy can be improved by cycle consistency loss~\cite{Reda_2019_ICCV}.
With additional image synthesis modules, TOF~\cite{xue2019video}, CtxSyn~\cite{Niklaus_2018_CVPR}, and BMBC~\cite{Park_2020_ECCV} improve the warped frame.
IM-Net~\cite{Peleg_2019_CVPR} achieves faster speed by considering multi-scale block-level horizontal/vertical motions.

Toward more sophisticated motion modeling, several novel approaches~\cite{Jiang_2018_CVPR, Yuan_2019_CVPR, Bao_2019_CVPR, Niklaus_2020_CVPR}, as well as higher-order representations, are proposed.
Quadratic~\cite{xu2019quadratic,qvi_iccvw19} and cubic~\cite{Chi_2020_ECCV} flows are estimated from multiple input frames.
On the other side, several methods introduce kernel-based modeling~\cite{Niklaus_2017_CVPR, Niklaus_2018_CVPR} to flow-based formulations.
DAIN~\cite{Bao_2019_CVPR} and MEMC-Net~\cite{MEMC-Net} implement adaptive warping by integrating the kernel and optical flow.
AdaCof~\cite{Lee_2020_CVPR} further unifies the combined representation in a similar formulation as deformable convolution~\cite{dai2017deformable}.
Using patch recurrency across spatial and time dimensions, Zuckerman~\etal~\cite{zuckerman2020across} have constructed a temporal SR model from a single video.

\Paragraph{Spatio-temporal super-resolution.}
A straightforward approach for STSR problem is to sequentially apply VSR and frame interpolation to a given video.
While this formulation handles spatial and temporal information separately, earlier methods~\cite{shechtman2002increasing, shechtman2005space} have solved the joint problem of STSR by optimizing very large objective terms.
Mudenagudi~\etal~\cite{mudenagudi2010space} adopt graph-cut, and Li~\etal~\cite{li2015space} utilize group cuts prior to deal with the problem.
Recently, STARnet~\cite{haris2020space} has proposed an end-to-end learnable framework for STSR problem using deep CNNs.
Xiang~\etal~\cite{xiang2020zooming} introduce a deformable ConvLSTM to effectively deal with the highly ill-posed problem of estimating smoother HR frames from a given low-frame-rate LR sequence.
In Track 2 of NTIRE 2021 Video Super-Resolution Challenge, we encourage participants to develop state-of-the-art models for the challenging STSR task.
\section{The Challenge}
\label{sec:challenge}
We have hosted NTIRE 2021 Video Super-Resolution Challenge to encourage participants to develop state-of-the-art VSR and STSR methods.
Following the previous NTIRE 2020 Challenge on Video Deblurring~\cite{Nah_2020_CVPR_Workshops_Deblur} and AIM 2019/2020 Challenge on Video Temporal Super-Resolution~\cite{Nah_2019_ICCV_Workshops_VTSR, Nah_2020_ECCV_Workshops_VTSR}, we adopt the REDS~\cite{Nah_2019_CVPR_Workshops_REDS} dataset to provide large-scale video frames for training and evaluation.

\subsection{REDS Dataset}
The REDS~\cite{Nah_2019_CVPR_Workshops_REDS} dataset contains 24,000, 3,000, and 3,000 HD-quality ($1280 \times 720$) video frames for training, evaluation, and test, respectively.
Each sequence in the dataset consists of 100 consecutive frames, e.g., \emph{`00000000.png'}--\emph{`00000099.png,'} that are sampled in 24fps.
From the ground-truth HR examples, we synthesize two datasets for the following challenge tracks.

\begin{table*}[t!]
    \centering
    \subfloat[Track 1. Spatial SR]{
        \begin{tabularx}{0.495\linewidth}{l >{\centering\arraybackslash}X >{\centering\arraybackslash}X >{\centering\arraybackslash}X >{\centering\arraybackslash}r}
            \toprule
            Team & PSNR$^\uparrow$ & SSIM$^\uparrow$ & LPIPS$_\downarrow$ & Runtime \\
            \midrule
            \textbf{NTU-SLab} & \textbf{33.36} & \textbf{0.9218} & \textbf{0.1115} & 7.3 \\
            Imagination & 32.96 & 0.9164 & 0.1247 & 7.5 \\
            model & 32.67 & 0.9121 & 0.1345 & 8.7 \\
            Noah\_Hisilicon\_SR & 32.65 & 0.9116 & 0.1358 & 22.6 \\
            VUE & 32.45 & 0.9085 & 0.1425 & 29.1 \\
            NERCMS & 32.13 & 0.9025 & 0.1491 & 2.2 \\
            Diggers & 31.97 & 0.9008 & 0.1565 & 1.2 \\
            \textit{withdrawn team} & 31.92 & 0.9010 & 0.1464 & 6.2 \\   
            MT.Demacia & 31.81 & 0.8914 & 0.1703 & 19.3 \\
            MiG\_CLEAR & 30.80 & 0.8774 & 0.1869 & 0.2 \\
            VCL\_super\_resolution & 30.66 & 0.8746 & 0.1898 & 0.5 \\
            SEU\_SR & 30.58 & 0.8722 & 0.1927 & 2.5 \\
            CNN & 29.67 & 0.8477 & 0.2164 & 1.4 \\
            Darambit & 28.65 & 0.8223 & 0.2621 & 13.8 \\
            \midrule
            bicubic upsampling & 26.48 & 0.7505 & 0.4393 & -\\
            \bottomrule
        \end{tabularx}
    }
    \subfloat[Track 2. Spatio-Temporal SR]{
        \begin{tabularx}{0.495\linewidth}{l >{\centering\arraybackslash}X >{\centering\arraybackslash}X >{\centering\arraybackslash}X >{\centering\arraybackslash}r}
            \toprule
            Team & PSNR$^\uparrow$ & SSIM$^\uparrow$ & LPIPS$_\downarrow$ & Runtime \\
            \midrule
            \textbf{Imagination} & \textbf{27.68} & \textbf{0.7772} & 0.2703 & 12.0 \\
            VUE & 27.39 & 0.7681 & 0.3230 & 20.2 \\
            TheLastWaltz & 27.01 & 0.7680 & 0.2777 & 2.1 \\
            sVSRFI & 26.92 & 0.7590 & 0.3231 & 5.2 \\
            T955 & 26.81 & 0.7617 & 0.2959 & 6.9 \\
            BOE-IOT-AIBD & 26.59 & 0.7570 & \textbf{0.2683} & 3.3 \\
            NaiveVSR & 26.46 & 0.7504 & 0.2967 & 2.8 \\
            VIDAR & 26.32 & 0.7613 & 0.3186 & 0.3 \\
            DeepBlueAI & 26.06 & 0.7413 & 0.3312 & 2.7 \\
            Team Horizon & 25.77 & 0.7341 & 0.3282 & 0.3 \\
            MiGMaster\_XDU & 25.75 & 0.7335 & 0.3367 & 2.0 \\
            superbeam & 25.61 & 0.7259 & 0.2893 & 2.2 \\
            CNN & 25.58 & 0.7284 & 0.3242 & 1.3 \\
            DSST & 24.92 & 0.7052 & 0.3813 & 2.5 \\
            \midrule
            upsample \& overlay & 23.11 & 0.6393 & 0.4978 & -\\
            \bottomrule
        \end{tabularx}
    }
    \tabcspace
    \caption{
        \textbf{NTIRE 2021 Video Super-Resolution Challenge results measured on the REDS~\cite{Nah_2019_CVPR_Workshops_REDS} test dataset.}
        Teams are ordered by ranks in terms of PSNR(dB).
        The running time is the average test time~(sec) taken to generate a single output image in reproduction process using 1 Quadro RTX 8000 GPU with 48GB VRAM.
        We note that the reported timing includes I/O and initialization overhead due to the difficulty in measuring pure model inference time by modifying each implementation.
    }
    \label{tab:result}
    \tabspace
\end{table*}

\Paragraph{Track1: Video Super-Resolution.}
We synthesize input $\times 4$ LR frames using MATLAB \code{imresize} function.
In other words, the conventional bicubic interpolation is used to generate the LR-HR pairs similar to existing SISR and VSR methods~\cite{Lim_2017_CVPR_Workshops, wang2019edvr}.
The goal of Track 1 is to reconstruct the ground-truth (GT) HR videos from the LR sequences.

\Paragraph{Track2: Video Spatio-Temporal Super-Resolution.}
From LR videos in Track 1, we remove odd-numbered frames, e.g., \emph{`00000001.png,'} \emph{`00000003.png,'} and so on, to formulate the STSR problem.
Therefore, the input sequences have $\times 2$ lower frame rate (12fps) than the original videos.
The goal of Track 2 is to perform STSR and reconstruct GT 24fps videos from the LR inputs.
Since the dataset for Track 2 is a subset of Track 1, all participants are requested not to use the odd-numbered frames from Track 1 when training their Track 2 methods.

\subsection{Metric and Evaluation}
We adopt two standard metrics to evaluate the submitted methods: PSNR and SSIM~\cite{wang2004image}.
Teams are sorted by the PSNR scores to determine the winner.
For reference, we also provide the LPIPS~\cite{zhang2018unreasonable} score to quantitatively measure the perceptual quality of result images.
The score is a type of distance function between reconstructed and GT frames defined on learned feature space.
We note that the LPIPS score is not considered for final ranking, but results with lower LPIPS tend to show better visual quality.

\section{Challenge Results}
\label{sec:results}
In the NTIRE 2021 Image Super-Resolution Challenge, each track has 247 and 223 registered participants, respectively.
During the final testing phase, 14 teams have submitted their solutions both for Track 1 and 2.
All teams are required to include full test frames, reproducible code, and corresponding fact sheets.
Table~\ref{tab:result} demonstrates the result of the challenge sorted by PSNR values.
To compare efficiency of the submitted solutions, we also measure runtime of each methods by executing the attached source code under the same environment.

\begin{figure*}[t!]
    \renewcommand{\wp}{0.13}
    \centering
    \subfloat{\includegraphics[width=\wp\linewidth]{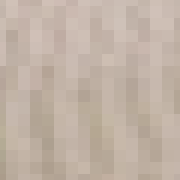}}
    \hfill
    \subfloat{\includegraphics[width=\wp\linewidth]{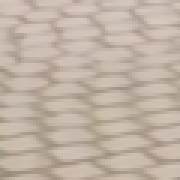}}
    \hfill
    \subfloat{\includegraphics[width=\wp\linewidth]{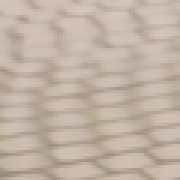}}
    \hfill
    \subfloat{\includegraphics[width=\wp\linewidth]{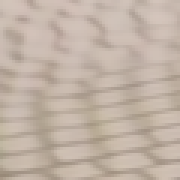}}
    \hfill
    \subfloat{\includegraphics[width=\wp\linewidth]{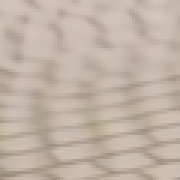}}
    \hfill
    \subfloat{\includegraphics[width=\wp\linewidth]{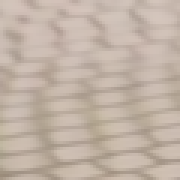}}
    \hfill
    \subfloat{\includegraphics[width=\wp\linewidth]{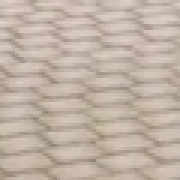}}
    \\
    \vspace{-3mm}
    \addtocounter{subfigure}{-7}
    \subfloat[Input]{\includegraphics[width=\wp\linewidth]{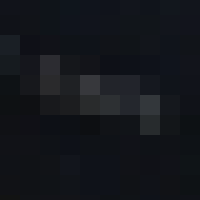}}
    \hfill
    \subfloat[]{\includegraphics[width=\wp\linewidth]{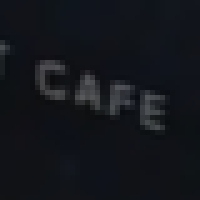}}
    \hfill
    \subfloat[]{\includegraphics[width=\wp\linewidth]{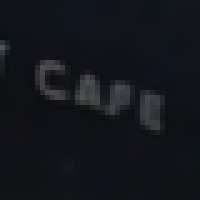}}
    \hfill
    \subfloat[]{\includegraphics[width=\wp\linewidth]{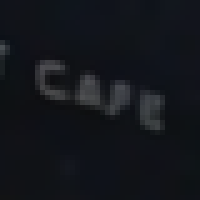}}
    \hfill
    \subfloat[]{\includegraphics[width=\wp\linewidth]{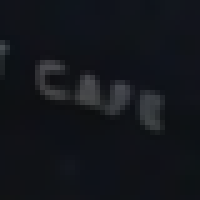}}
    \hfill
    \subfloat[]{\includegraphics[width=\wp\linewidth]{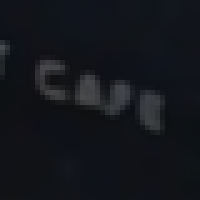}}
    \hfill
    \subfloat[GT]{\includegraphics[width=\wp\linewidth]{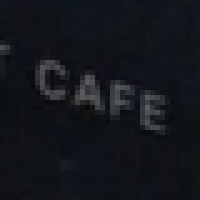}}
    \\
    \figcspace
    \caption{
        \textbf{Comparison between top-ranked results in Track 1.}
        (b) NTU-SLab team.
        (c) Imagination team.
        (d) model team.
        (e) Noah\_Hisilicon\_SR team.
        (f) VUE team.
        Patches are cropped from REDS (test) `\emph{006/00000097.png}' and `\emph{008/00000099.png},' respectively.
    }
    \label{fig:comp_track1}
    \figspace
\end{figure*}
\begin{figure*}[t!]
    \renewcommand{\wp}{0.13}
    \centering
    \subfloat{\includegraphics[width=\wp\linewidth]{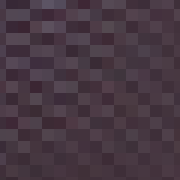}}
    \hfill
    \subfloat{\includegraphics[width=\wp\linewidth]{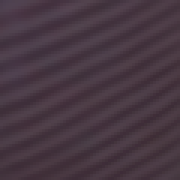}}
    \hfill
    \subfloat{\includegraphics[width=\wp\linewidth]{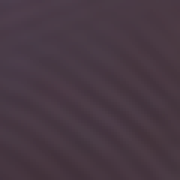}}
    \hfill
    \subfloat{\includegraphics[width=\wp\linewidth]{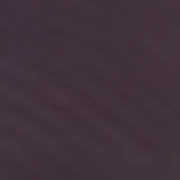}}
    \hfill
    \subfloat{\includegraphics[width=\wp\linewidth]{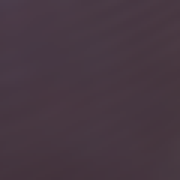}}
    \hfill
    \subfloat{\includegraphics[width=\wp\linewidth]{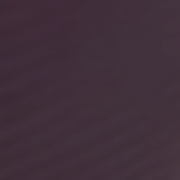}}
    \hfill
    \subfloat{\includegraphics[width=\wp\linewidth]{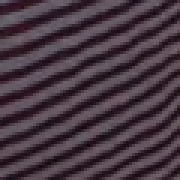}}
    \\
    \vspace{-3mm}
    \addtocounter{subfigure}{-7}
    \subfloat[Input]{\includegraphics[width=\wp\linewidth]{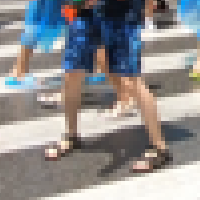}}
    \hfill
    \subfloat[]{\includegraphics[width=\wp\linewidth]{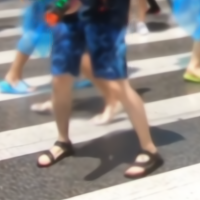}}
    \hfill
    \subfloat[]{\includegraphics[width=\wp\linewidth]{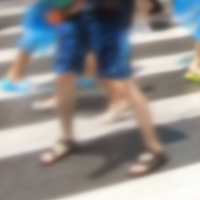}}
    \hfill
    \subfloat[]{\includegraphics[width=\wp\linewidth]{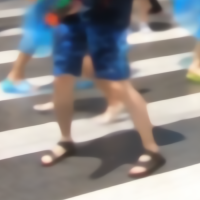}}
    \hfill
    \subfloat[]{\includegraphics[width=\wp\linewidth]{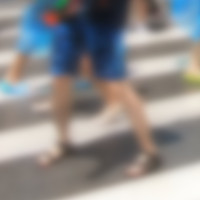}}
    \hfill
    \subfloat[]{\includegraphics[width=\wp\linewidth]{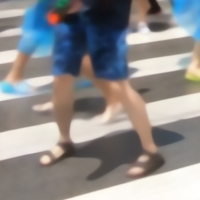}}
    \hfill
    \subfloat[GT]{\includegraphics[width=\wp\linewidth]{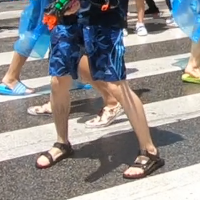}}
    \\
    \figcspace
    \caption{
        \textbf{Comparison between top-ranked results in Track 2.}
        (b) Imagination team.
        (c) VUE team.
        (d) TheLastWaltz team.
        (e) sVSRFI team.
        (f) T955 team.
        We note that the inputs are provided for timesteps $t - 1$ and $t + 1$ only, while the output frame corresponds to the timestep $t$.
        For simplicity, we only visualize the input at $t - 1$.
        Patches are cropped from REDS (test) `\emph{006/00000001.png}' and `\emph{013/00000009.png},' respectively.
    }
    \label{fig:comp_track2}
    \figspace
\end{figure*}

\subsection{Challenge winners}
In Track 1: Video Super-Resolution, NTU-SLab team got the first place with BasicVSR++ architecture.
The major advance of BasicVSR++ from the baseline BasicVSR is a novel second-order grid propagation strategy.
Please check Section~\ref{ssec:ntu} for more detail.

In Track 2: Video Spatio-Temporal Super-Reoslution, Imagination team got the first place with a combination of Local to Context Video Super-Resolution (LCVR) and Multi-scale Quadratic Video Interpolation (MQVI).
Specifically, the multi-scale interpolation method is used to refine the odd frames that are not provided as inputs.
Please refer to Section~\ref{ssec:imagination} for more explanation.

\subsection{Visual comparison}
In this section, we provide a visual comparison between top-ranked teams.
Figure~\ref{fig:comp_track1} shows super-resolved outputs from an input LR REDS image.
Figure~\ref{fig:comp_track2} illustrates HR interpolated frames from several methods, where only neighboring LR frames are given to the STSR models.

\section{Challenge Methods and Teams}
\label{sec:methods}
In this section, we describe the submitted solution and details based on the challenge fact sheets.

\subsection{NTU-SLab}
\label{ssec:ntu}

\begin{figure}[h]
	\centering
    \includegraphics[width=\linewidth]{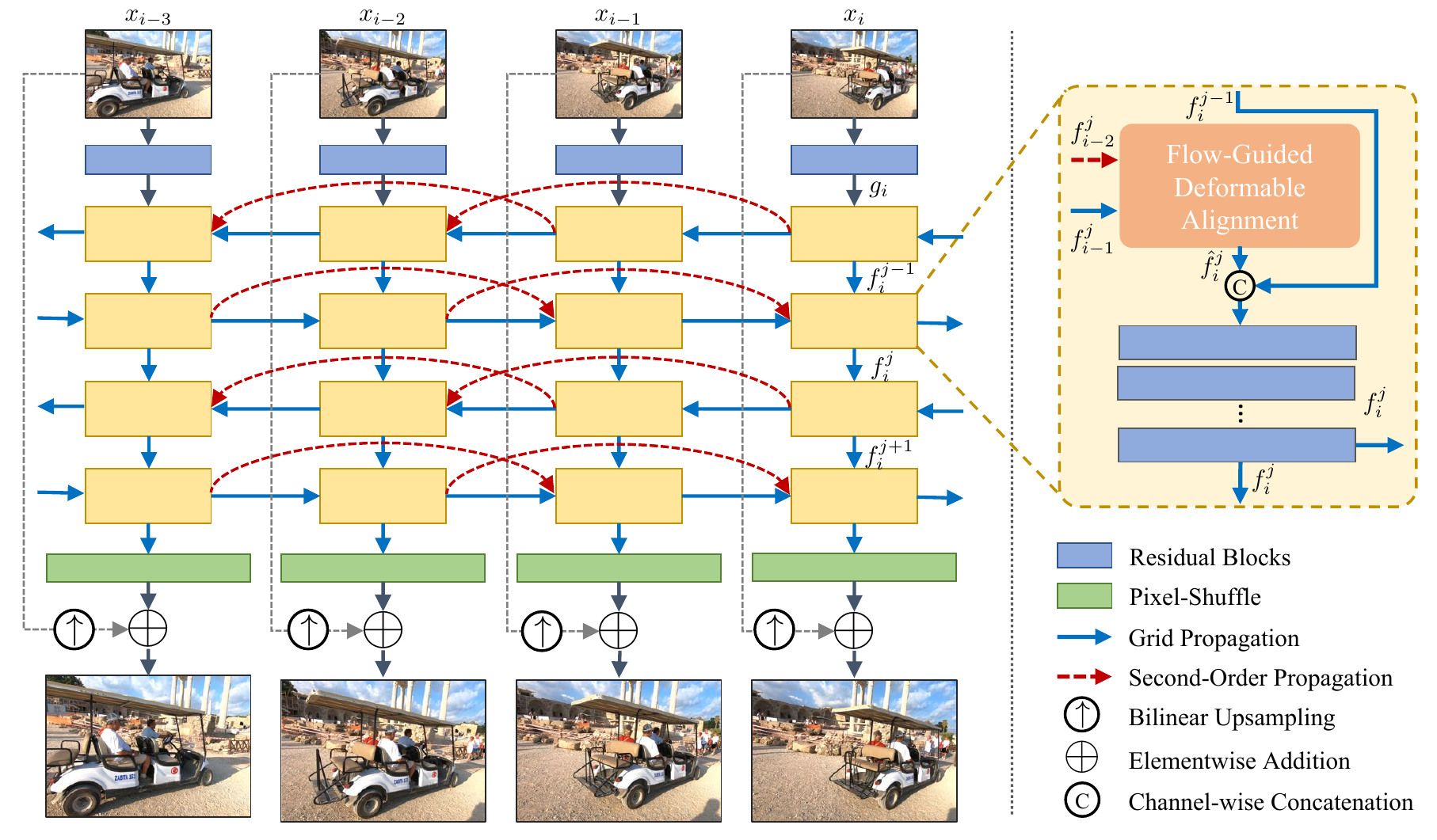}
    \\
    \figcspace
    \caption{
        \textbf{NTU-SLab team (Track 1)}. BasicVSR++
    }
	\label{fig:ntu_track1}
	\figspace
\end{figure}

\begin{figure}[h]
	\centering
    \includegraphics[width=\linewidth]{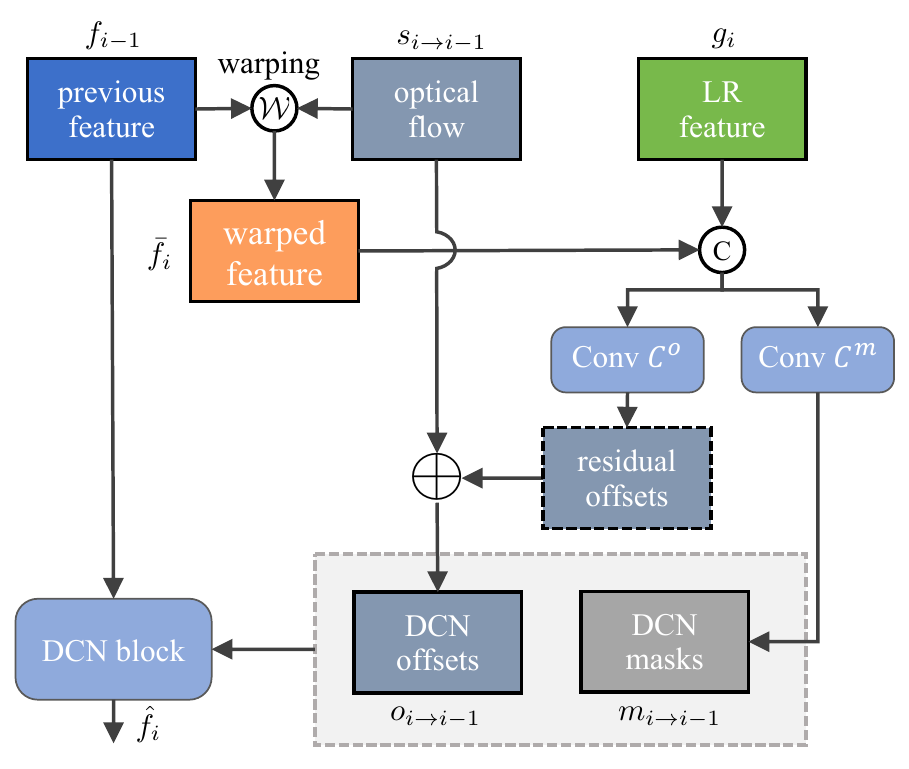}
    \\
    \figcspace
    \caption{
        The flow-guided deformable alignment module from NTU-SLab Team.
    }
	\label{fig:ntu_track2}
	\figspace
\end{figure}

NTU-SLab team proposes BasicVSR++ framework which is an enhanced version of BasicVSR~\cite{chan2020basicvsr}.
Following the same basic methodology as BasicVSR, the propagation and alignment methods are modified by introducing second-order grid propagation and flow-guided deformable alignment.
Figure~\ref{fig:ntu_track1} shows the overall architecture of BasicVSR++.
The feature sequences are extracted from the image using residual blocks, and they are propagated under the second-order grid propagation scheme.
Inside the propagation block, the flow-guided deformable alignment module is added for better performance.

To overcome the limit that the features can be propagated only once in BasicVSR, the second-order grid propagtaion scheme enables the features to be refined multiple times.
Through the multiple bidirectional propagation layers, the features of different time steps are revisited to make each feature include more useful information.
The second-order connection (red dotted lines in Figure~\ref{fig:ntu_track1}), the information is more aggregated from different spatio-temporal locations.

Flow-guided deformable alignment uses deformable convolution instead of flow-based alignment.
Deformable convolution shows better performance than flow-based alignment as deformable convolution has offset diversity.
They use optical flow to alleviate the training instability of deformable convolution.
The architecture of the flow-guided deformable alignment is shown in Figure~\ref{fig:ntu_track2}.

\subsection{Imagination team}
\label{ssec:imagination}
\begin{figure}[h]
	\centering
	\subfloat[Track 1]{\includegraphics[width=\linewidth]{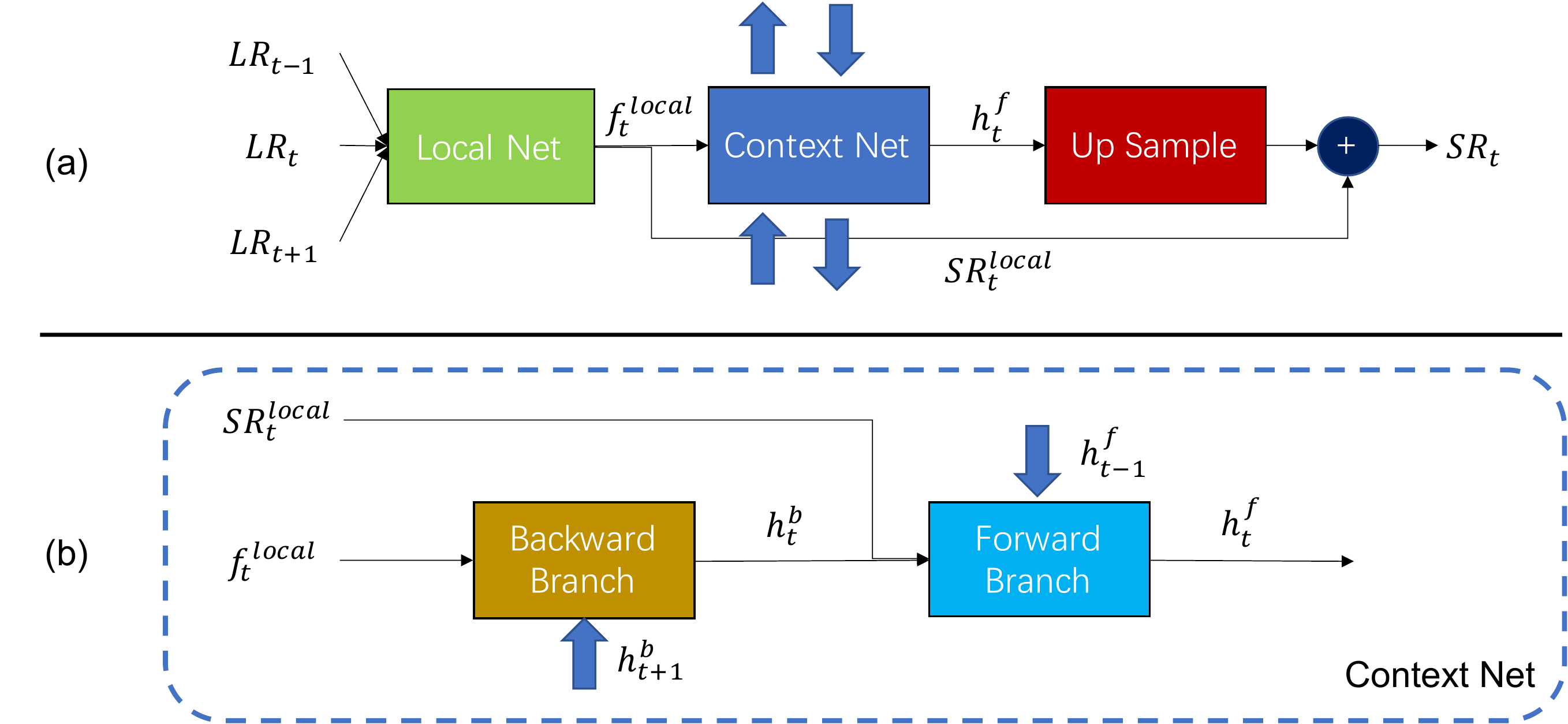}}
	\\
	\vspace{-3mm}
	\subfloat[Track 2]{\includegraphics[width=\linewidth]{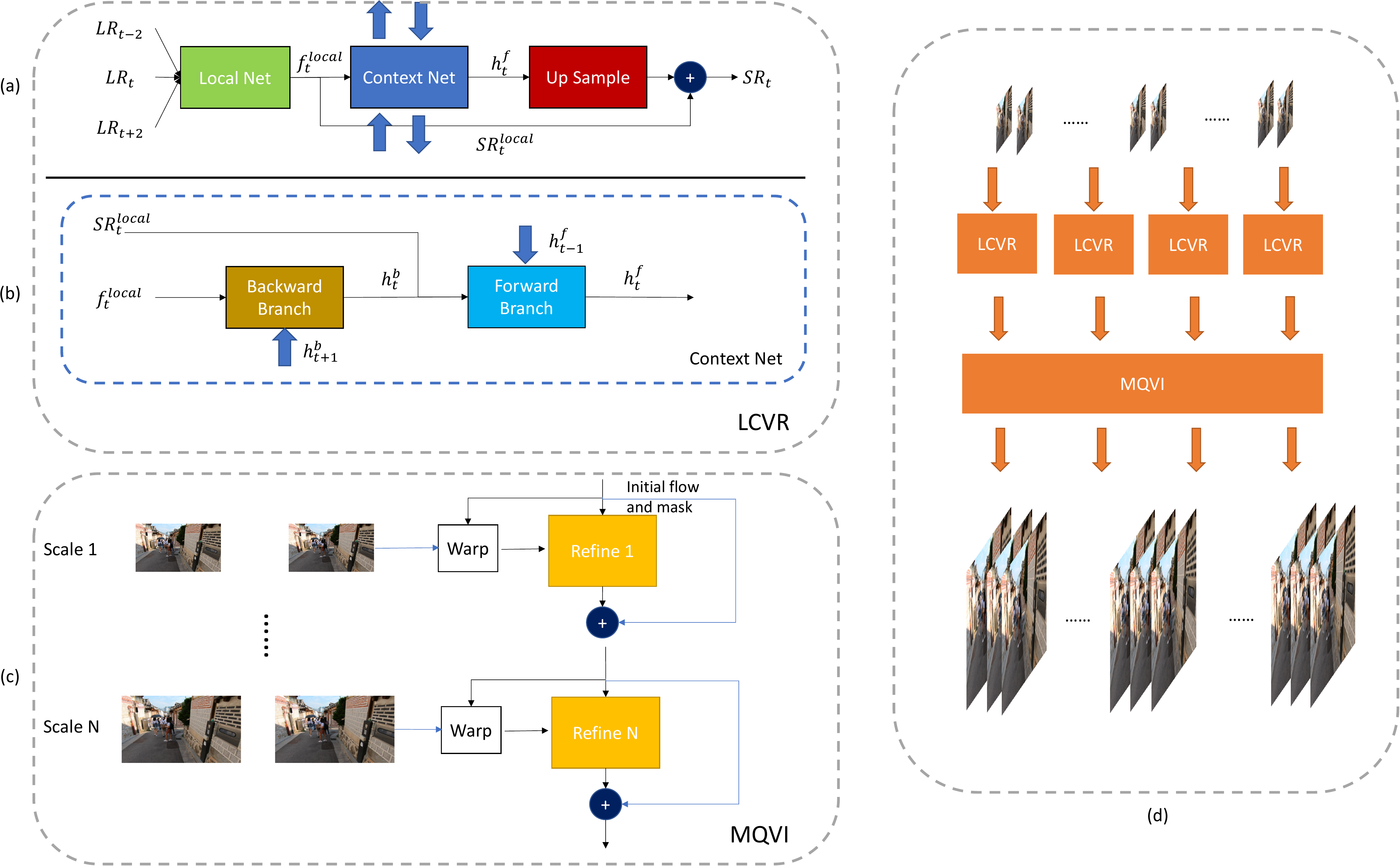}}
    \\
    \figcspace
    \caption{
        \textbf{Imagination team (Track 1 \& 2)}. Local to Context (Multi-level Quadratic model for) Video Super-Resolution
    }
	\label{fig:imagination_track1}
	\figspace
\end{figure}

Imagination team proposes a Local to Context Video Super-Resolution, LCVR in short, to conduct video super-resolution, and combined it with Multi-scale Quadratic Video Interpolation, MQVI, to conduct video spatial-temporal super-resolution.

LCVR framework consists of three modules: local net module, context net module, and upsample module.
The EDVR network~\cite{wang2019edvr} with channel attention is used for the local net module.
The local feature and the super-resolved frame are generated from the local net module.
The context net module is composed of backward and forward branch and its output is converted to the frame residual by passing through the upsample module.
The final SR result is made by summing the SR frame output of the local net module and the frame residual estimated by the context net module and the upsample module.
The additional self-ensemble strategy is used to boost the performance by 0.2 dB.
The overview of LCVR framework is shown in (a) of Figure~\ref{fig:imagination_track1}.

In the video spatio-temporal super-resolution track, the MQVI module is attached after the LCVR module.
At first, the HR-size images of even frame indices are generated by the LCVR module and the odd frames are interpolated using MQVI module.
They adopt Quadratic frame interpolation since it can handle more complex motion than linear frame interpolation.
In addition, the multi-scale structure applied to refine the feature in coarse-to-fine manner.
The MQVI and overall structure is illustrated in (b) of Figure~\ref{fig:imagination_track1}.

\subsection{model}

\begin{figure}[h]
	\centering
    \includegraphics[width=\linewidth]{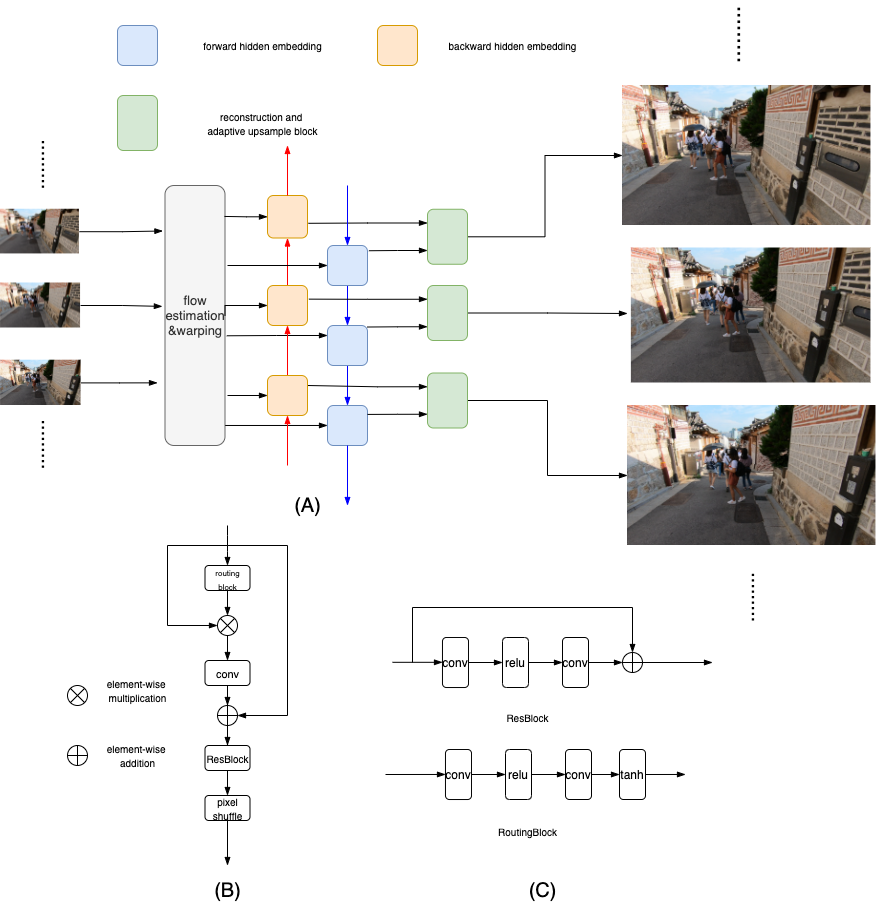}
    \\
    \figcspace
    \caption{
        \textbf{model team (Track 1)}. Flow-Alignment + Bi-directional Encoding + Adaptive Up-sampling
    }
	\label{fig:model_track1}
	\figspace
\end{figure}

model team suggests a framework combining flow alignment module, bidirectional encoding module and adaptive upsampling module.
They use flow estimation from SpyNet~\cite{ranjan2017optical} to warp features.
The concatenated and warped features are put into bidirectional encoding layers, loading some useful context information from other timesteps.
Finally the adaptive upsampling module reconstructs the final output using routing block.
Routing block plays role as the spatial attention of the feature.
In Figure~\ref{fig:model_track1}, the overview of the model is shown in (A), and the adaptive upsampling module is shown in (B), (C).

\subsection{Noah\_Hisilicon\_SR}

Noah\_Hisilicon\_SR team proposes LGFFN, Local and Global Feature Fusion Network.
Based on BasicVSR framework, they improve it in two aspects.
First, they combine the global propagation feature and local propagation feature.
Second, they incorporate unsupervised learning scheme into flow estimation module for better performance.

\begin{figure}[h]
	\centering
    \includegraphics[width=\linewidth]{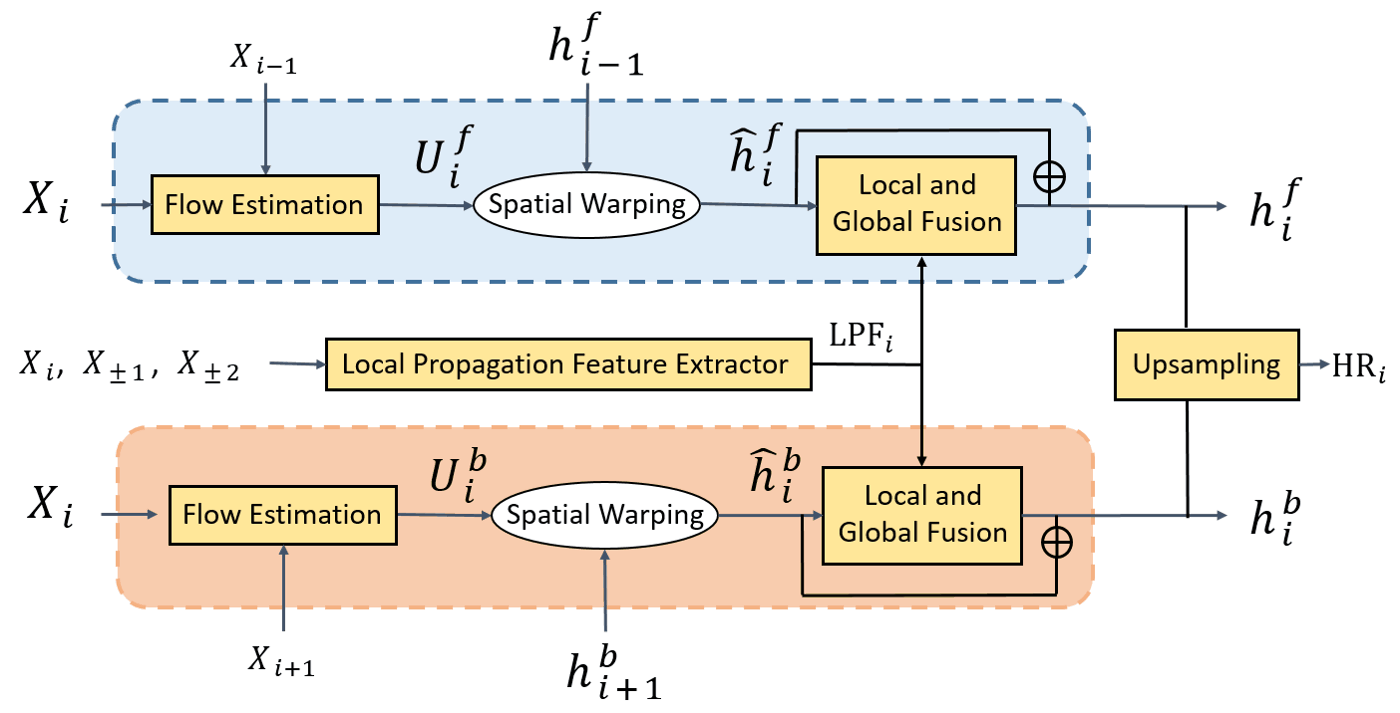}
    \\
    \figcspace
    \caption{
        \textbf{Noah\_Hisilicon\_SR team (Track 1)}. Local and Global Feature Fusion Network
    }
	\label{fig:noah_track1}
	\figspace
\end{figure}

Figure~\ref{fig:noah_track1} describes the feature fusion architecture of proposed framework.
The blue and the red box indicates the forward and the backward propagation, respectively.
While the two blocks imply the global propagation, the local propagation feature extractor estimates the local feature.
The two features are combined by local and global fusion block for better reconstruction performance.

There is also an improvement in flow estimation module.
Since the previous SR methods bring flow estimation module which is trained using synthesized flow dataset, the pretrained module usually suffers from the discrepancy between the synthesized flow dataset and REDS dataset.
To resolve the issue, they adopt the unsupervised scheme like to train the flow estimation module in REDS dataset directly.
The two images in 3 adjacent frames are put into the network and warped to each other using the estimated optical flow.
To be specific, given the two image $I_1, I_2$, the network estimates the forward and the backward flow $U_{12}, U{21}$ and the estimated warped frame is generated as $\hat{I}_{1}(p) = I_2(p + U_{12}(p))$, where $p$ is a pixel of the image.
Then the model is trained using the distance between the original image and the warped image from the other frame.

\subsection{VUE}

\begin{figure}[h]
	\centering
	\subfloat[Track 1]{\includegraphics[width=\linewidth]{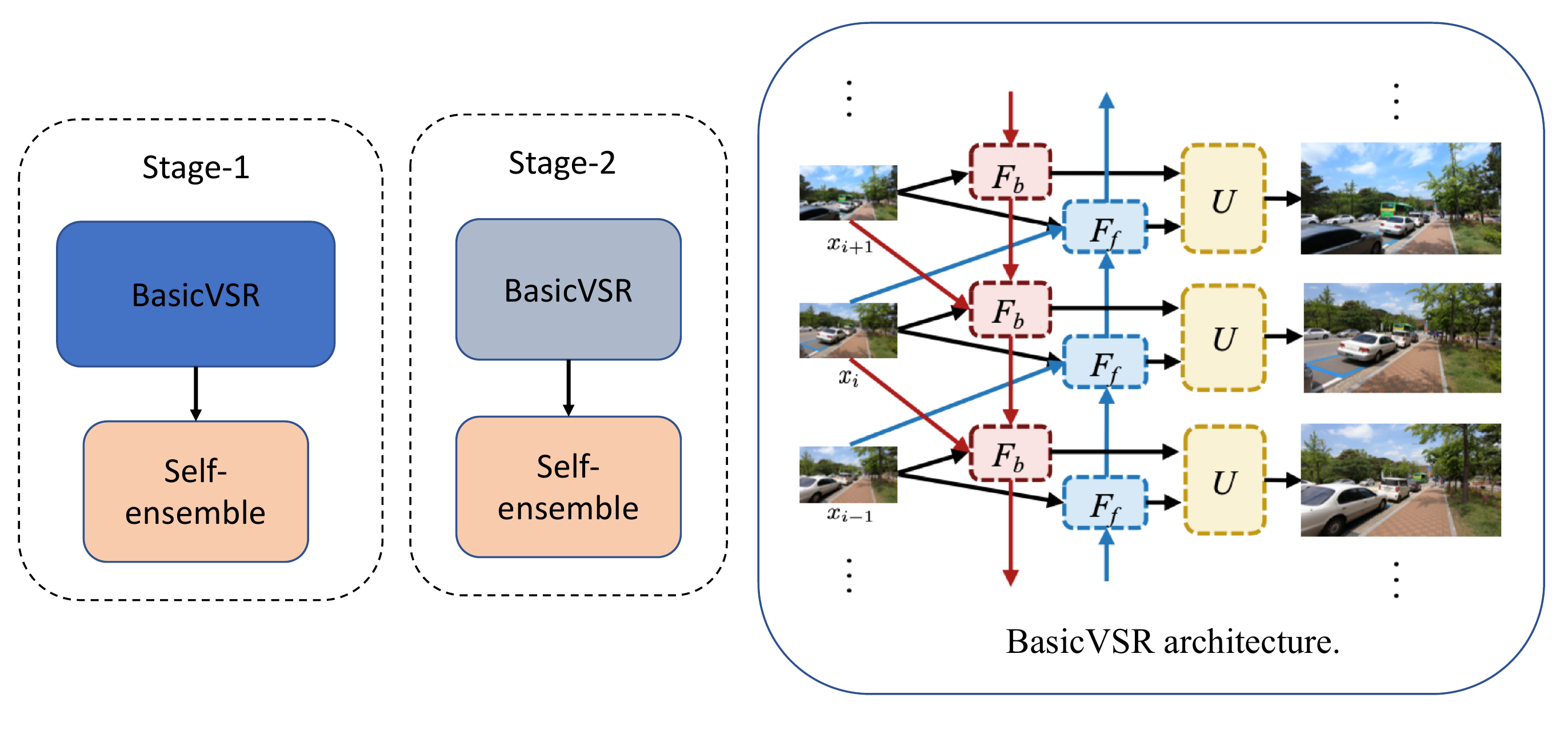}}
	\\
	\vspace{-3mm}
    \subfloat[Track 2]{\includegraphics[width=\linewidth]{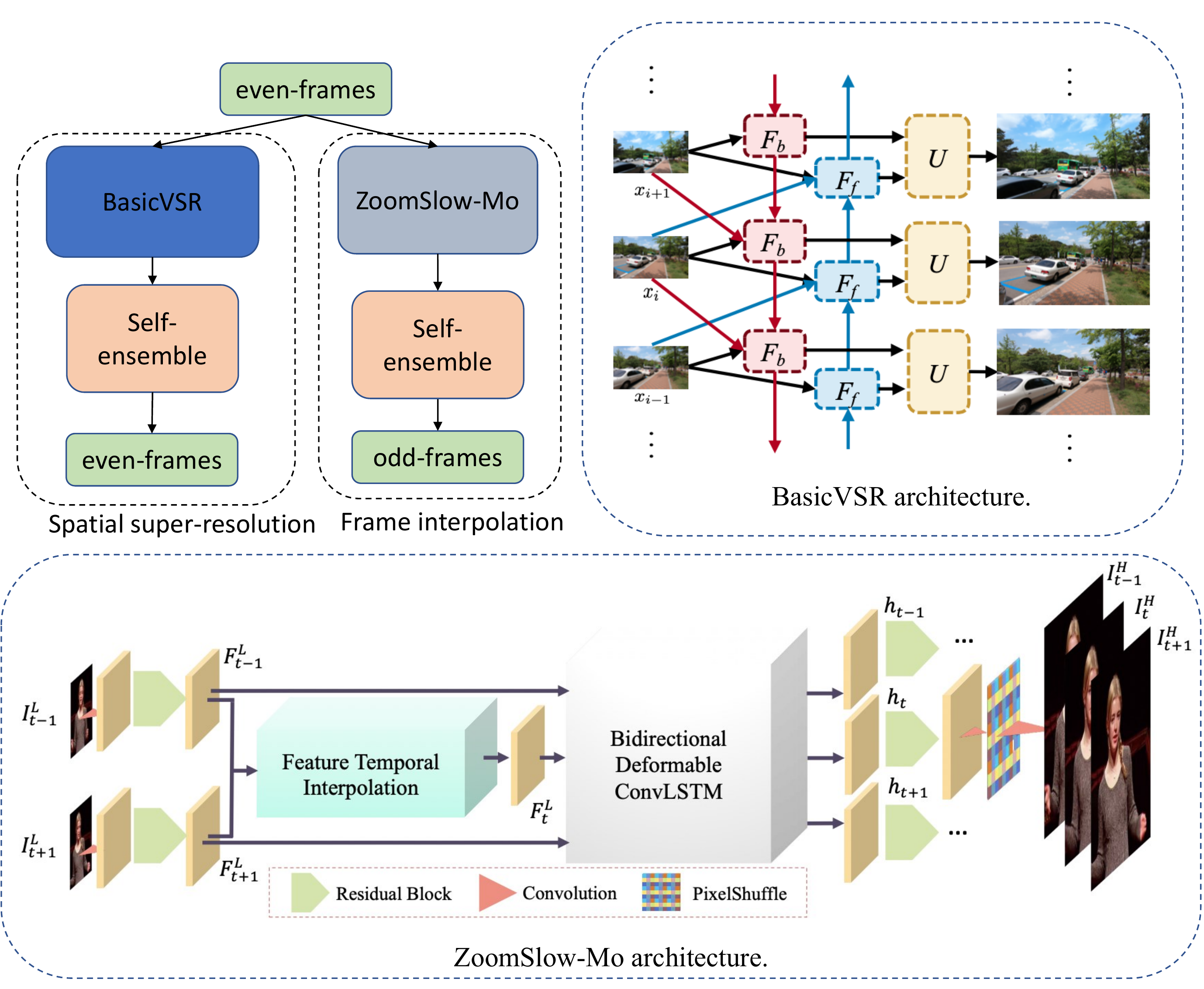}}
    \\
    \figcspace
    \caption{
        \textbf{VUE team (Track 1 \& 2)}. Two-Stage BasicVSR framework
    }
	\label{fig:vue_track1}
	\figspace
\end{figure}

VUE team proposes two-stage algorithms in both tracks as shown in Figure~\ref{fig:vue_track1}.
In track 1, BasicVSR~\cite{chan2020basicvsr} model is used in both stages, and the output of the first stage is fed into the second stage.
The coarse super-resolved images are generated at first, and they are refined by passing through the second network.
In both stages, the self-ensemble strategy is used for better performance.

In track 2, the two stages are run in parallel and the second stage is replaced with Zooming Slow-mo~\cite{xiang2020zooming} model.
The BasicVSR model estimates the super-resolved output of even frames, and Zooming Slow-mo model estimates the super-resolved output of odd frames.
Self-ensemble is also applied to both stages for higher PSNR.

\subsection{NERCMS}

NERCMS team proposes omniscient video super-resolution network (OVSR), which is composed of two sub-networks: precursor network and successor network.
The precursor network scans the LR input frames and generate SR frames and hidden states of all time steps, and the successor network refines the SR frames using the two outputs of precursor network.
They use progressive fusion residual blocks to build both networks.

\subsection{Diggers}

\begin{figure}[h]
	\centering
    \includegraphics[width=\linewidth]{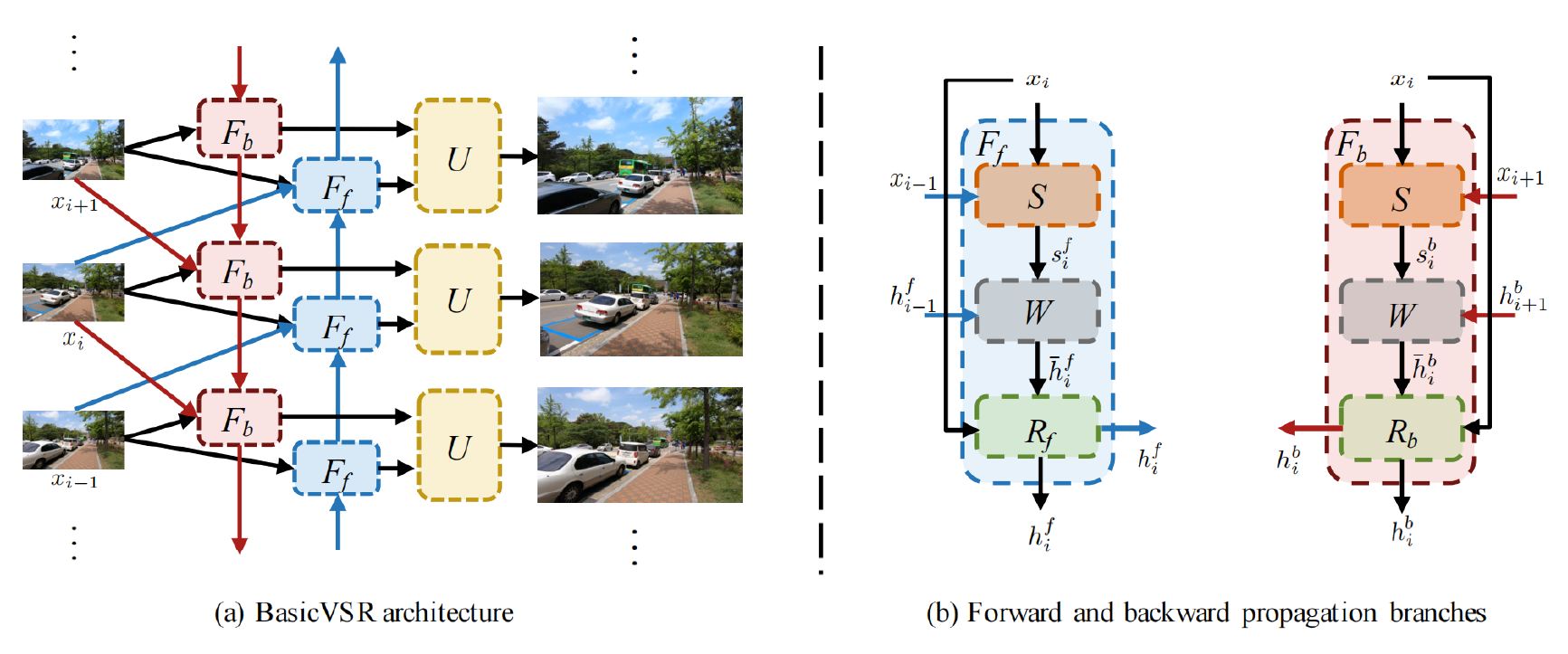}
    \\
    \figcspace
    \caption{
        \textbf{Diggers team (Track 1)}. basicVSR
    }
	\label{fig:diggers_track1}
	\figspace
\end{figure}

Diggers team reproduces BasicVSR~\cite{chan2020basicvsr} network as shown in Figure~\ref{fig:diggers_track1}.

\subsection{MT.Demacia}

The proposed framework consists of two stages.
At the first stage, the context information of temporally adjacent frames are aggregated using the PCD and TSA module proposed in EDVR~\cite{wang2019edvr}.
In order to explore the useful information and remove the redundant information, the non-local block is added to each frame, separately.
Finally, the stacked channel-attention residual block is applied for the reconstruction.
The second stage is image SR stage, and the details of the image is added from the output of the first stage.

\subsection{MiG\_CLEAR}

MiG\_CLEAR team improves EDVR~\cite{wang2019edvr} architecture in two ways.
They adopt self-calibrated convolution at the PCD module of EDVR network.
The self-calibrated convolution can help estimating the offset better in deformable convolution.
There is also an improvement in TSA module.
They replace TSA module to the integration of Temporal Group Attention (TGA) module and channel attention.
Since REDS dataset has large movements, computing the correlation of each time step might not be accurate.
The group attention module before computing the temporal attention can alleviate the issue.
Finally, the channel attention enable the model to achieve better performance.

\subsection{VCL\_super\_resolution}

\begin{figure}[h]
	\centering
    \includegraphics[width=\linewidth]{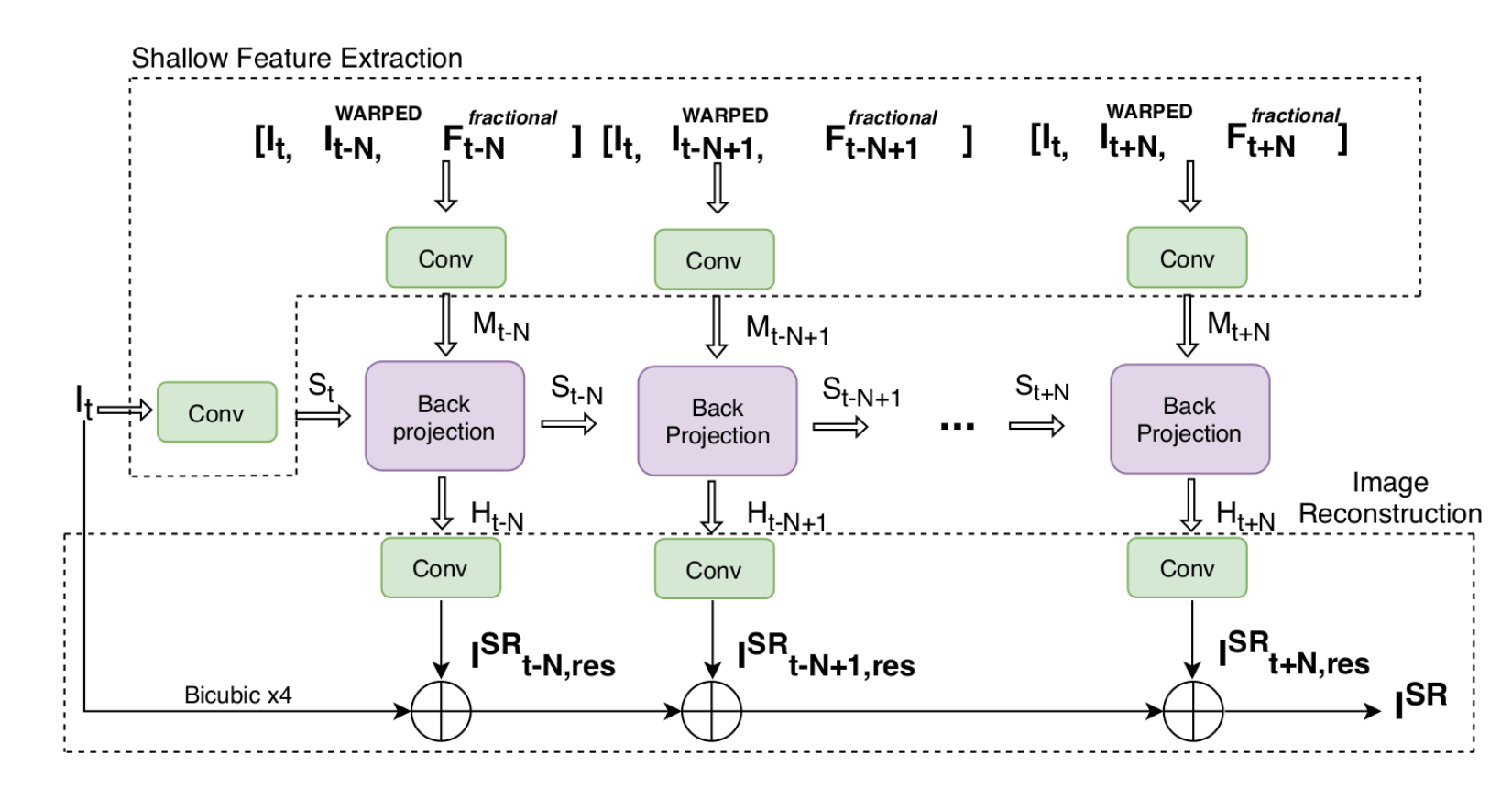}
    \\
    \figcspace
    \caption{
        \textbf{VCL\_super\_resolution team (Track 1)}. Quantized Warping and Residual Temporal Integration for Video Super-Resolution on Fast Motions
    }
	\label{fig:vcl_track1}
	\figspace
\end{figure}

The framework proposed by VCL\_super\_resolution team is similar to RBPN~\cite{haris2019recurrent} has two processes: motion compensation and super-resolution.
Using the characteristic of video super-resolution task, motion compensation process is a warping from the neighbor frame to the current frame using integer interpolation.
The fractional part of the displacement is not applied in warping, but it is provided to the next stage as input.
After the compensation of features, super-resolution is applied to make HR-size outputs.
The target image is generated by adding the extra information from neighbor frames using the back-projection process.
The shallow feature which is extracted from the LR input data is fed into the back-projection module.
After aggregating the information from other time steps in the module, the output at each cell is passed to the reconstruction module, and the super-resolved output is estimated.
The super-resolution process is shown in Figure~\ref{fig:vcl_track1}.

\subsection{SEU\_SR}

\begin{figure}[h]
	\centering
    \includegraphics[width=\linewidth]{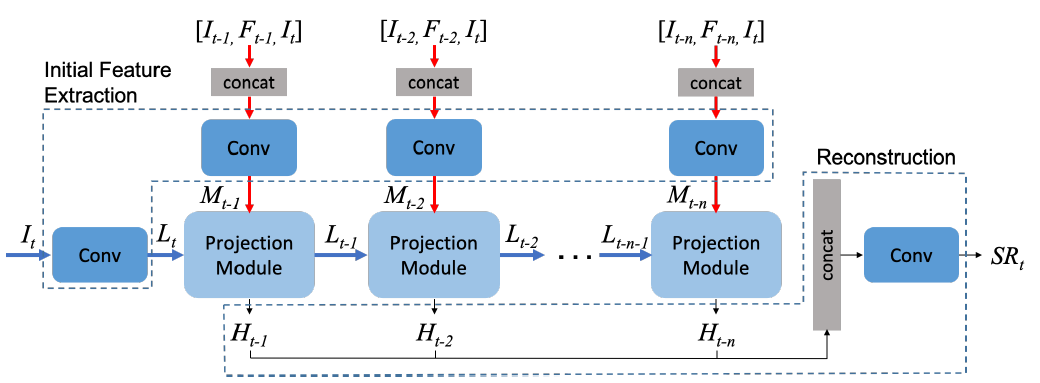}
    \\
    \figcspace
    \caption{
        \textbf{SEU\_SR team (Track 1)}. Recurrent Back-Projection Network
    }
	\label{fig:seu_track1}
	\figspace
\end{figure}

SEU\_SR team applies RBPN~\cite{haris2019recurrent} network to REDS dataset, and the architecture is depicted in Figure~\ref{fig:seu_track1}.
Using the input image and the flow maps from the other time steps, the projection module aggregates the information from different time steps.
The outputs of projection modules are put through the layers to generate the HR-size image at the current time step.

\subsection{CNN}

\begin{figure}[h]
\centering
	\subfloat[Track 1]{\includegraphics[width=\linewidth]{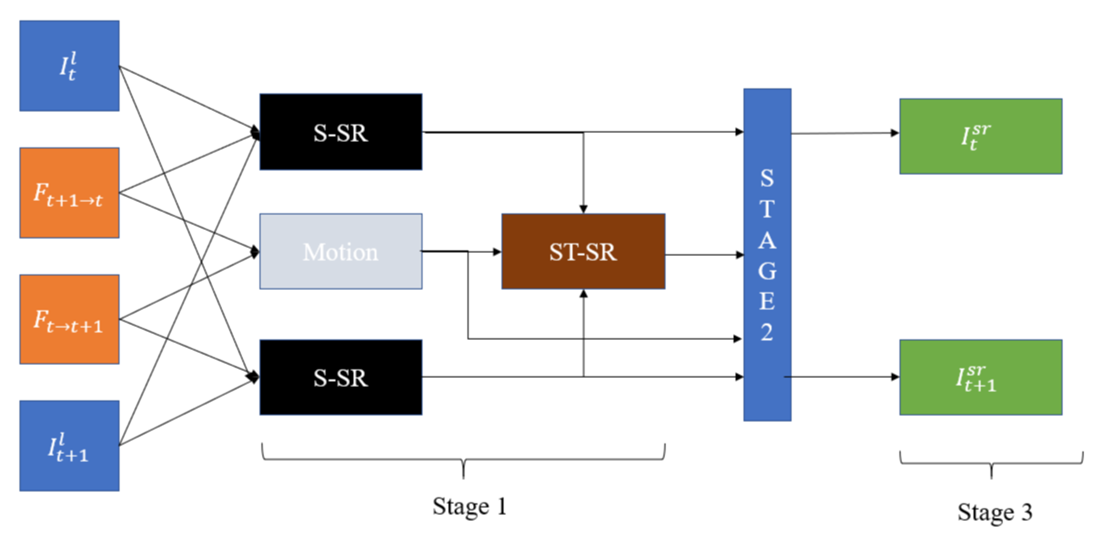}}
	\\
	\vspace{-3mm}
    \subfloat[Track 2]{\includegraphics[width=\linewidth]{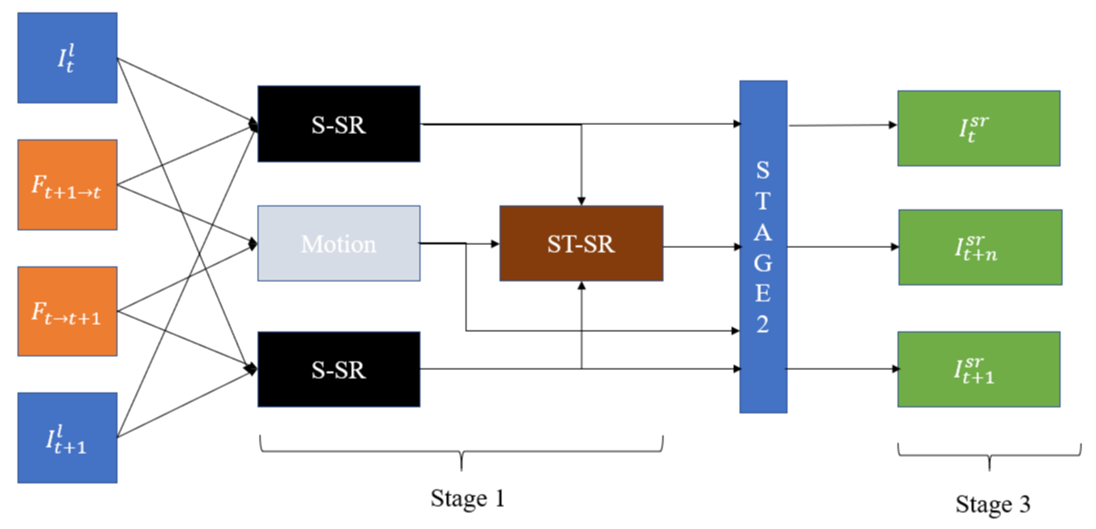}}
    \\
    \figcspace
    \caption{
        \textbf{CNN team (Track 1 \& 2)}. Space-Time-Aware Multi-Resolution Video Enhancement
    }
	\label{fig:cnn_track1}
	\figspace
\end{figure}

CNN team applies STARNet~\cite{haris2020space} model to REDS dataset in both tracks.
The top image of Figure~\ref{fig:cnn_track1} is the architecture used in track 1 and the bottom corresponds to track 2.

\subsection{Darambit}

\begin{figure}[h]
	\centering
    \includegraphics[width=\linewidth]{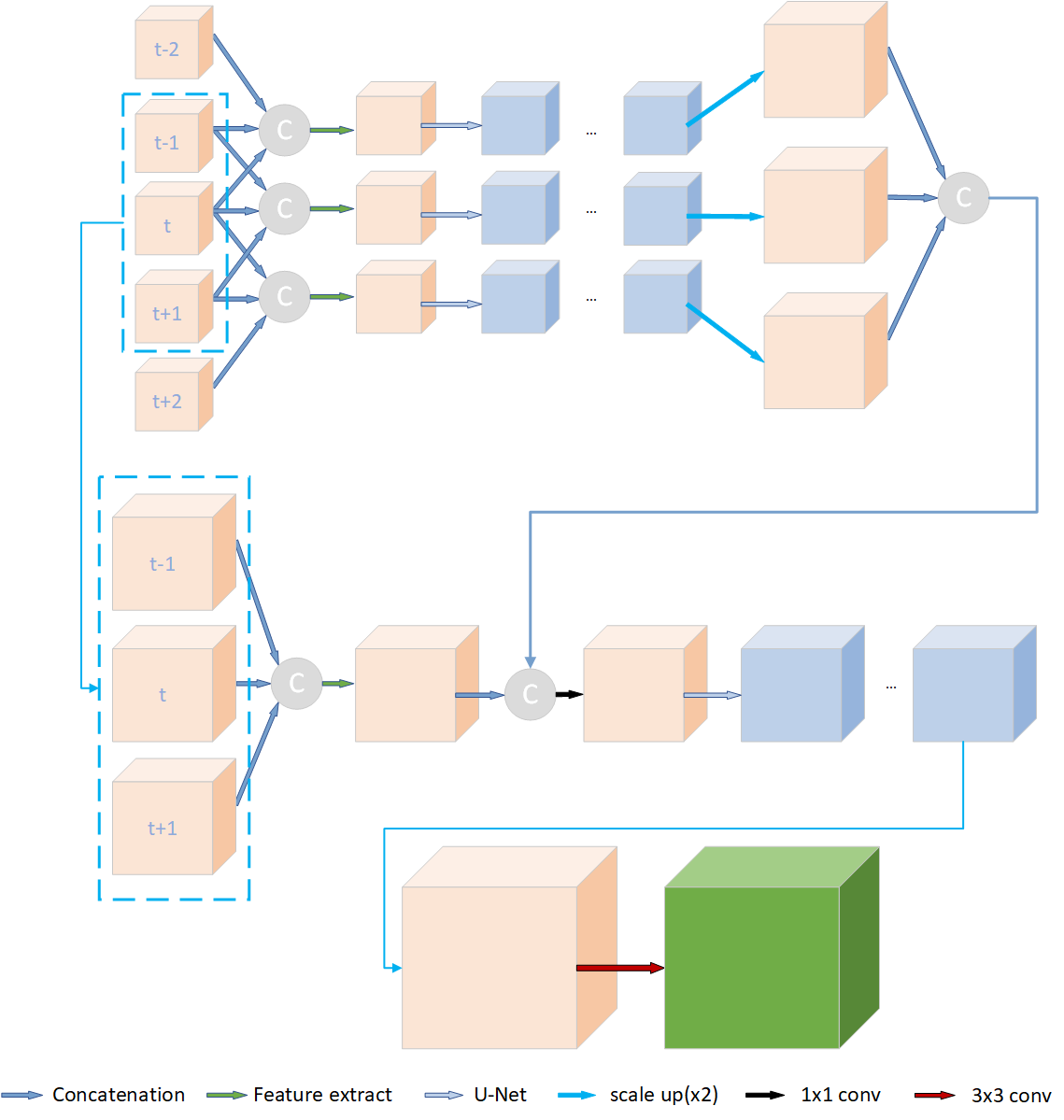}
    \\
    \figcspace
    \caption{
        \textbf{Darambit team (Track 1)}. Multi-frame Feature Combination Network for Video Super-Resolution
    }
	\label{fig:darambit_track1}
	\figspace
\end{figure}

Darambit team proposes U-net like architecture to estimate the target image.
Five consecutive frames are put into the network, and the frames at the center are concatenated to make the local features, and the features are propagated through the network to generate the final output.
The whole process is shown in Figure~\ref{fig:darambit_track1}.

\subsection{TheLastWaltz}

\begin{figure}[h]
	\subfloat[Overview]{\includegraphics[width=\linewidth]{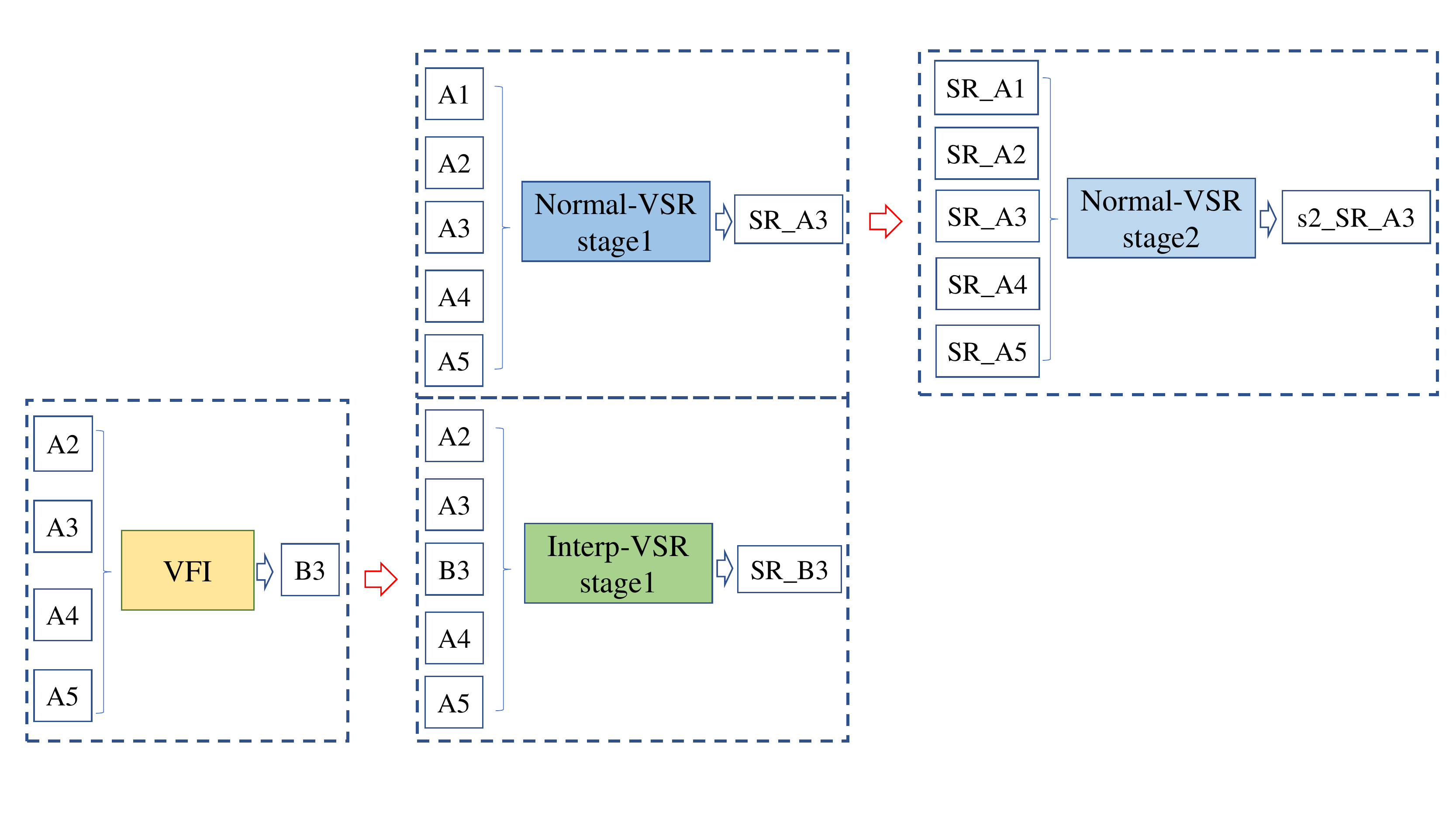}}
	\\
	\vspace{-2mm}
    \subfloat[Normal-VSR stage 2]{\includegraphics[width=\linewidth]{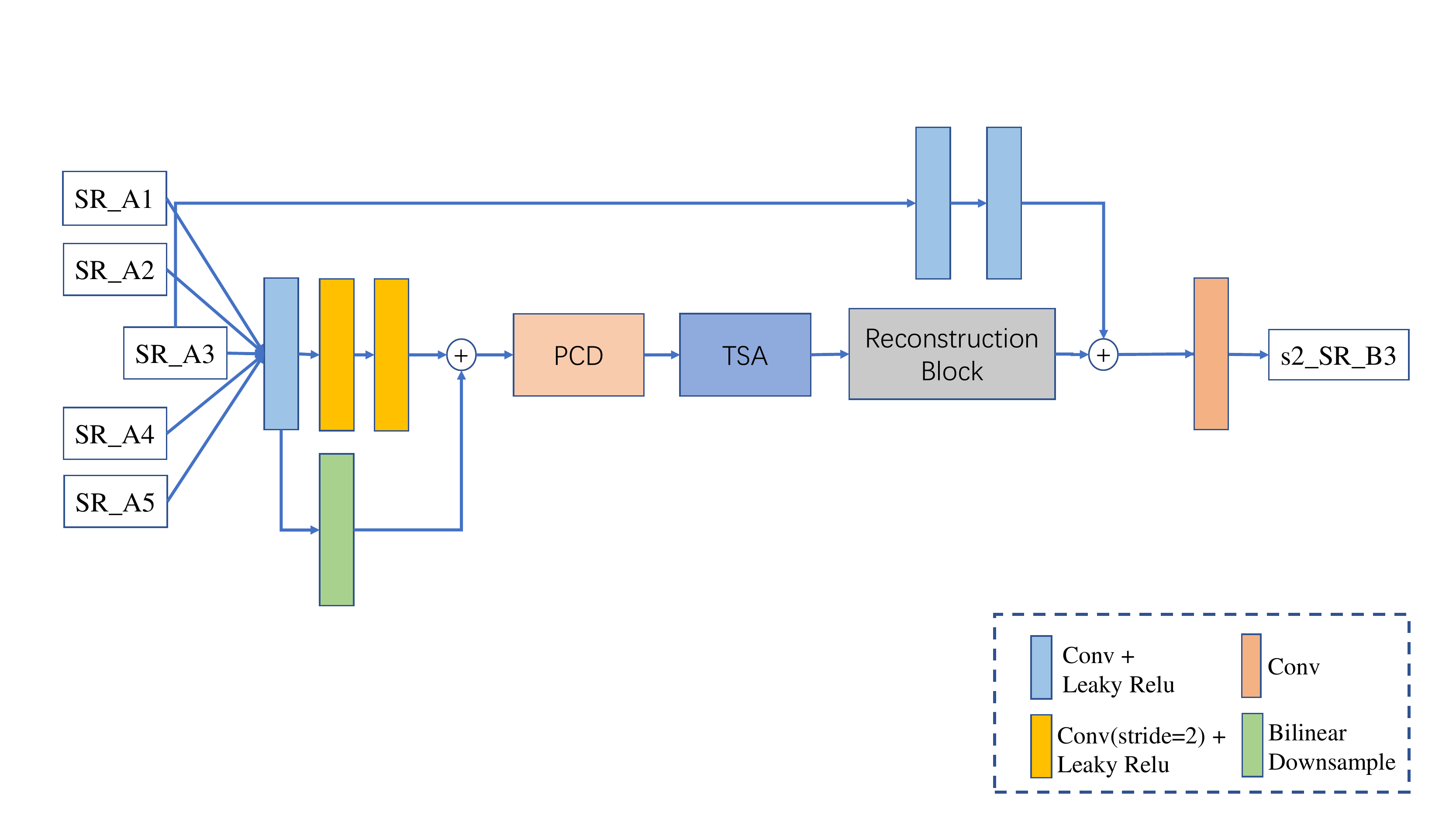}}
    \\
    \caption{
        \textbf{TheLastWaltz team (Track 2)}. Separated framework for spatio-temporal super-resolution
    }
	\label{fig:thelastwaltz_track2}
	\figspace
\end{figure}

TheLastWaltz team divide the video super-resolution task into three parts: video frame interpolation (VFI), normal frame super-resolution (Normal-VSR), and interpolated frame super-resolution (InterpSR).
The $A$s and $B$s in Figure~\ref{fig:thelastwaltz_track2} (a) are the frames arranged in following order: $A_0, B_0, A_1, B_1, ..., A_n, B_n$.
The $A$ frames are given even frames and $B$s are frames to be interpolated.
After LR-size odd frames are interpolated, the interpSR model estimates the super-resolved version of B, SR\_B.
When super-resolving the odd frames, four neighbor even frames are used to provide relevant information.
In other words, $A_{i-2}, A_{i-1}, A_{i+1}, A_{i+2}$ are used when making $B_{i}$.
They use quadratic video interpolation technique~\cite{xu2019quadratic} for the interpolation module.
Then, the interpolated frame and the inputs used for the interpolation are put into the interpSR model, generating the super-resolved of $B_i$.

The super-resolution of odd frames, denoted as normal VSR, is composed of two stages.
Similar to the two-stage training technique mentioned in EDVR~\cite{wang2019edvr}, the architecture of the second stage is modified.
Instead of directly add the concatenated input to the final summation, the input is processed by passing through two CNN layers.
The author argues that the input of the normal-VSR stage 2 is the output of the normal-VSR stage 1, which is sufficiently optimized by training the stage 1 model.
Thus, directly put the output to the network makes the optimization difficult, and the CNN layers can alleviate the problem.
The modified architecture of the normal-VSR stage 2 is illustrated in (b) of Figure~\ref{fig:thelastwaltz_track2}.

\begin{figure}[h]
	\centering
    \includegraphics[width=\linewidth]{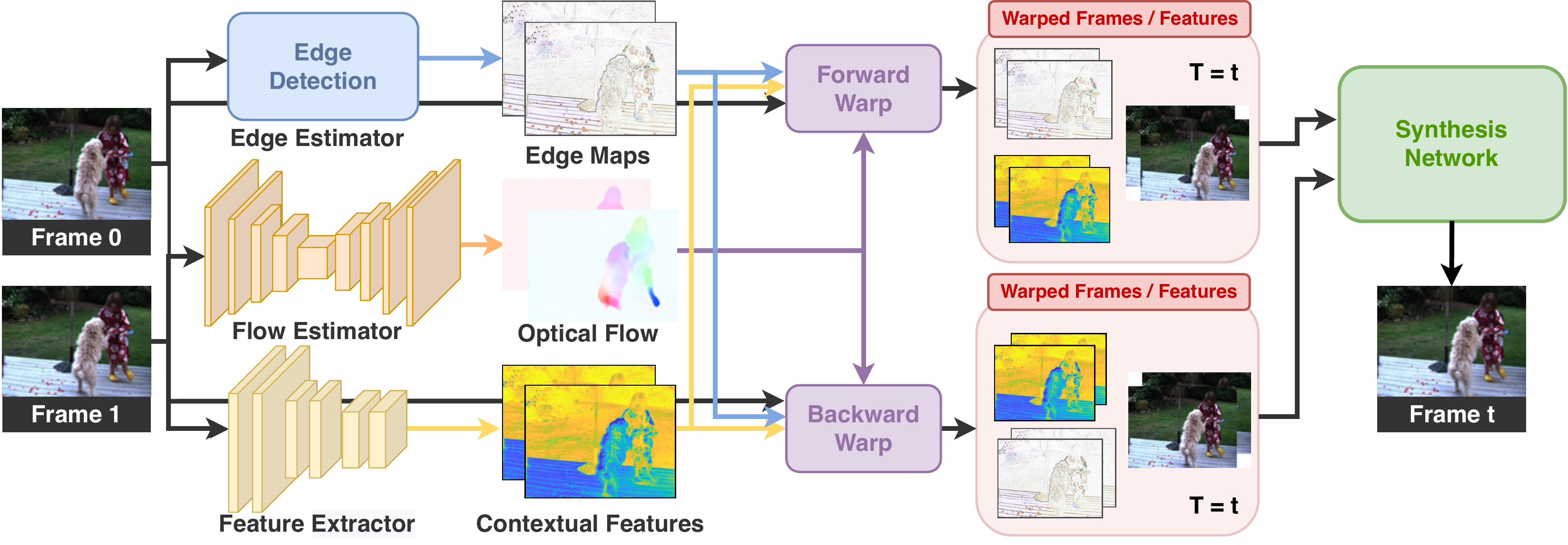}
    \\
    \figcspace
    \caption{
        \textbf{sVSRFI team (Track 2)}. A two-stage method for video spatio-temporal super-resoution
    }
	\label{fig:svsrfi_track2}
	\figspace
\end{figure}

\subsection{sVSRFI}

sVSRFI team proposes a framework, which is a combination of video super-resolution and video frame interpolation.
Contrary to the previous team, the super-resolution is executed before the frame interpolation.
BasicVSR model is used for super-resolving LR inputs.
They propose the bidirectional warping method for video frame interpolation.
Using the edge map, optical flow and contextual features are warped and the synthesis network generates the interpolated frames.
The video frame interpolation architecture is shown in Figure~\ref{fig:svsrfi_track2}.

\subsection{T955}

\begin{figure}[h]
	\centering
    \includegraphics[width=\linewidth]{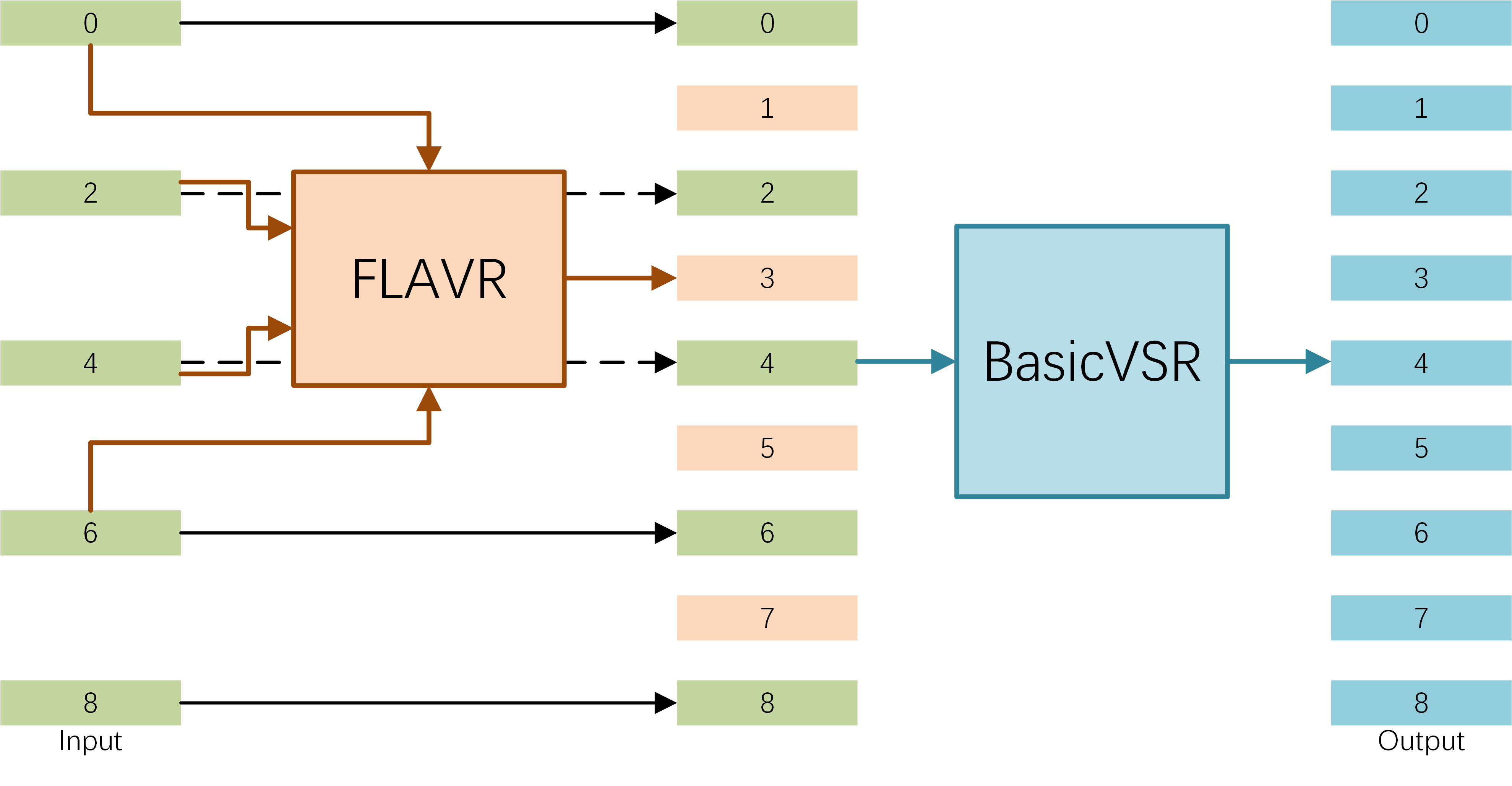}
    \\
    \figcspace
    \caption{
        \textbf{T955 team (Track 2)}. BasicVSR after FLAVR
    }
	\label{fig:t955_track2}
	\figspace
\end{figure}

T955 team proposes a combination of FLAVR~\cite{kalluri2020flavr} and BasicVSR~\cite{chan2020basicvsr}.
The odd frames are estimated by FLAVR, and the whole sequence is super-resolved by BasicVSR.
Figure~\ref{fig:t955_track2} describes the overall procedure.

\subsection{BOE-IOT-AIBD}

\begin{figure}[h]
	\centering
    \includegraphics[width=\linewidth]{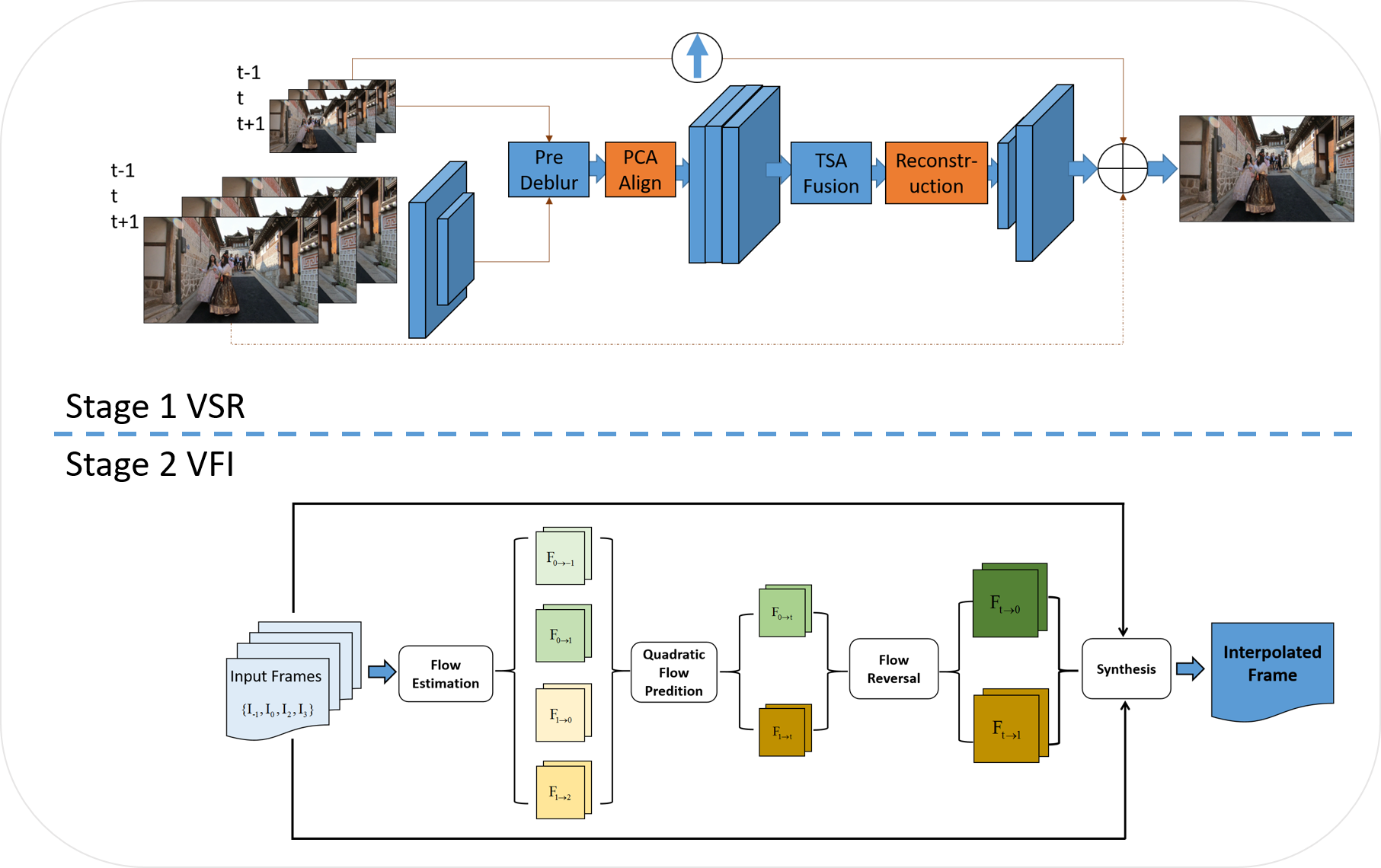}
    \\
    \figcspace
    \caption{
        \textbf{BOE-IOT-AIBD team (Track 2)}. Two-Stage Video Spatial \& Temporal Super-Resolution Algorithm
    }
	\label{fig:boe_track2}
	\figspace
\end{figure}

The framework of BOE-IOT-AIBD team consists of two stages, video super-resolution and temporal super-resolution network.
They adopt EDVR~\cite{wang2019edvr} as the video super-resolution and Multi Scale Quadratic Interpolation (MSQI)~\cite{Son_2020_ECCV_Workshops} approach as temporal super-resolution network
The illustration of two networks is shown in Figure~\ref{fig:boe_track2}.

\subsection{NaiveVSR}

\begin{figure}[h]
	\centering
    \includegraphics[width=\linewidth]{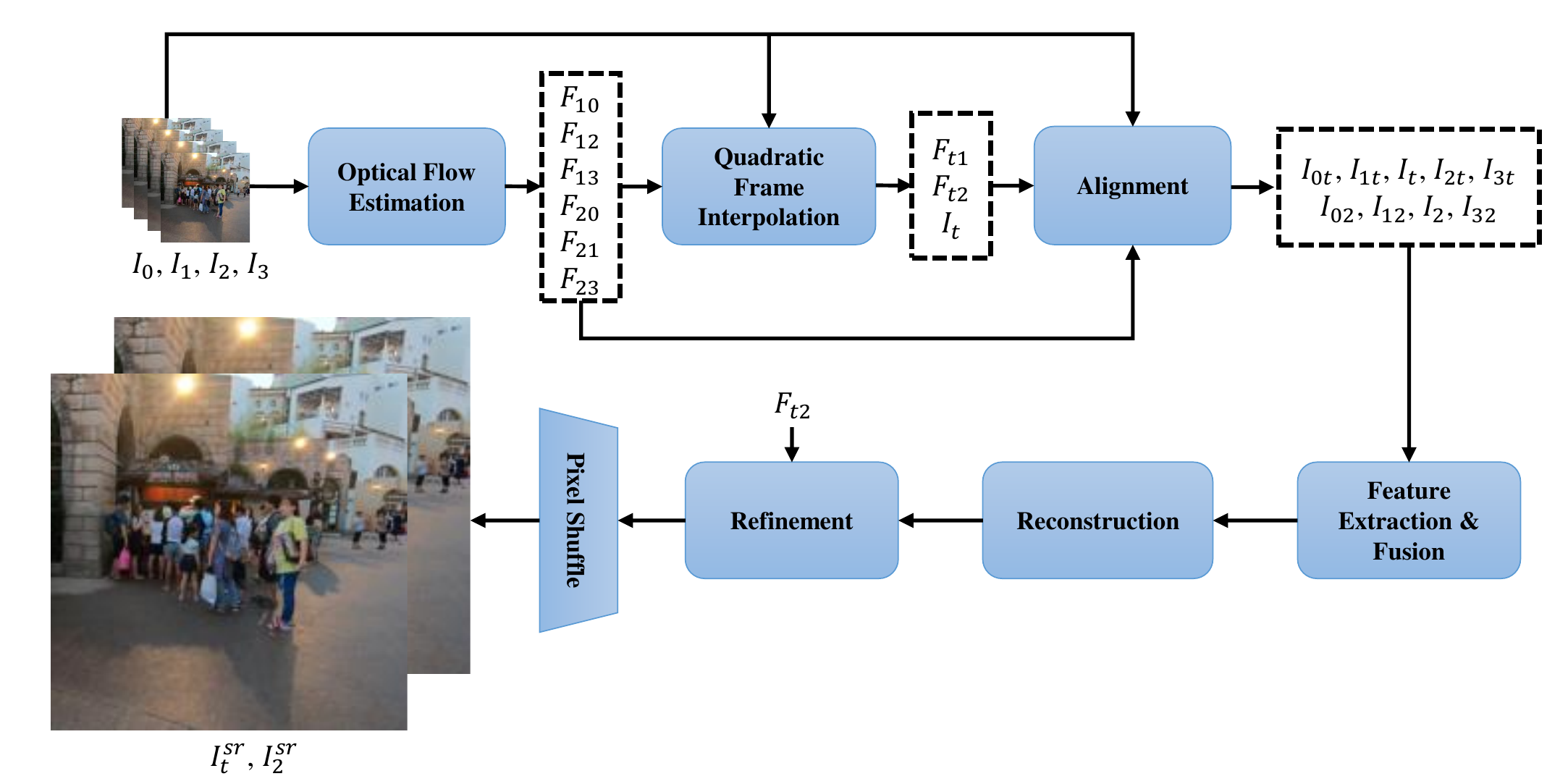}
    \\
    \figcspace
    \caption{
        \textbf{NaiveVSR team (Track 2)}. Quadratic Space-Time Video Super-Resolution
    }
	\label{fig:naivevsr_track2}
	\figspace
\end{figure}

NaiveVSR team proposes the network which executes the video frame interpolation first, followed by the video super-resolution.
Figure~\ref{fig:naivevsr_track2} shows the overview of their method.
$I_0 - I_3$ is four consecutive images among the given even frame low-resolution inputs and the interpolated time step, $t$, is set to 1.5 in this work.
They use Enhanced Quadratic video interpolation model~\cite{liu2020enhanced} for interpolation, but they remove residual contextual synthesis module in their method since the following super-resolution network can restore the detail of the interpolated image.
Using the 5 low-resolution images, $I_0, I_1, I_t, I_2, I_3$, the video-super resolution network generates the HR-size images of two frame indices ($I_t, I_2$).
First they warp the other frames to $I_t, I_2$, respectively, and the concatenated aligned inputs are put into the network.
They adopt TSA fusion module of EDVR~\cite{wang2019edvr} to obtain the feature at frame t, 2.
The features are put through the reconstruction module and the refinement module.
Before the refinement module, the two features are aligned and the aligned features are fed into the refinement module.
Finally, they upsample the feature to get the super-resolved images.
The overall process is shown in Figure~\ref{fig:naivevsr_track2}.

\subsection{VIDAR}

\begin{figure}[b]
	\centering
	\subfloat[Overall structure of the three-stage network.]{\includegraphics[width=\linewidth]{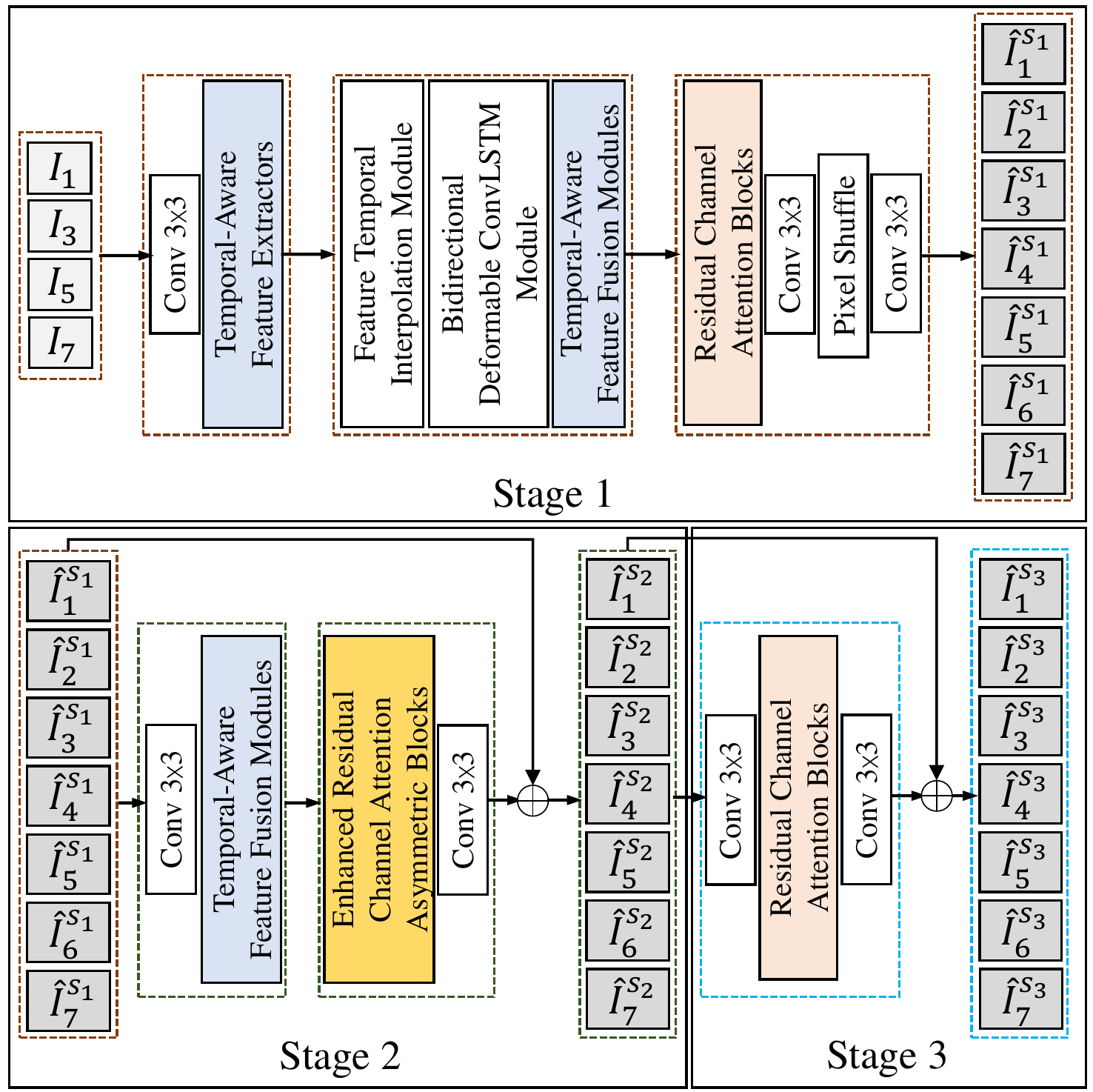}}
	\\
	\vspace{-3mm}
    \subfloat[Detailed structure of temporal-aware feature extractor (TAFE)]{\includegraphics[width=\linewidth]{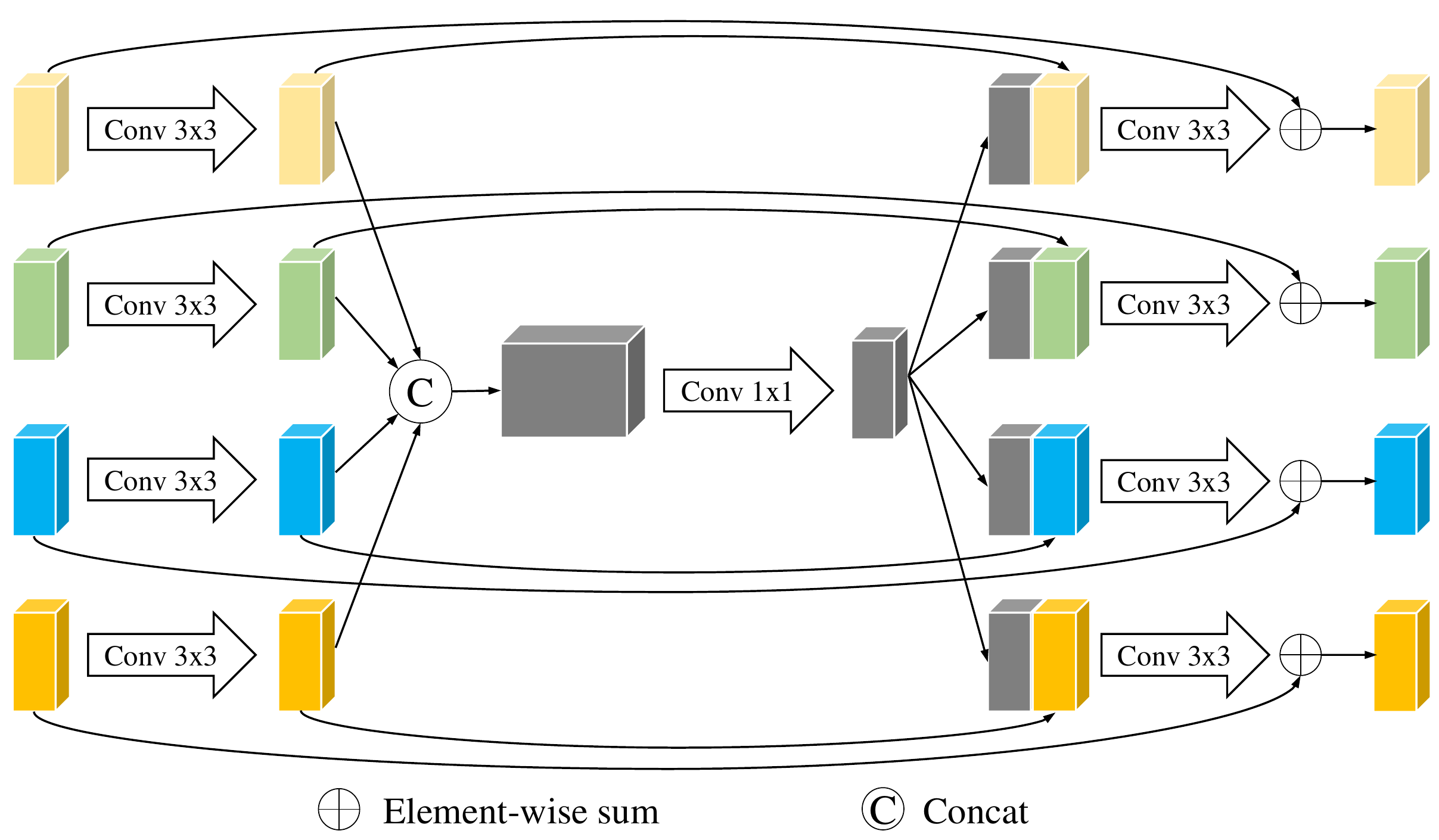}}
    \\
    \figcspace
    \caption{
        \textbf{VIDAR team (Track 2)}. Enhanced Temporal Alignment and Interpolation Network for Space-Time Video Super-Resolution
    }
	\label{fig:vidar_track2}
	\figspace
\end{figure}

VIDAR team proposes a three-stage network, one for joint video super-resolution and interpolation and two for optimizing and refining the outputs. 
The overall structure is shown in Figure~\ref{fig:vidar_track2} (a). 
$I_1$, $I_3$, $I_5$ and $I_7$ are the input frames. 
$\hat{I_i}^{s_1}$, $\hat{I_i}^{s_2}$ and $\hat{I_i}^{s_3}$ ($i \in [1, 7]$) are the output frames from stage 1, stage 2 and stage 3, respectively.
The core components of the network are the temporal-aware feature extractor (TAFE) and the temporal-aware feature fusion (TAFF) module. Figure~\ref{fig:vidar_track2} (b) shows the detailed structure of TAFE.
The only difference between TAFE and TAFF is the number of inputs: 4 are put into TAFE and 7 are put into TAFF.
To better exploit the temporal information, they adopt the temporal profile loss proposed in \cite{xiao2020space}.
In stage 1, they use 8 TAFEs, 3 TAFFs and 12 residual channel attention blocks~\cite{zhang2018image}.
They use 10 TAFFs and 30 enhanced residual channel attention asymmetric blocks~\cite{Liu_2020_CVPR_Workshops} in stage 2, and 30 residual channel attention blocks in stage 3.


\subsection{DeepBlueAI}

\begin{figure}[h]
	\centering
    \includegraphics[width=\linewidth]{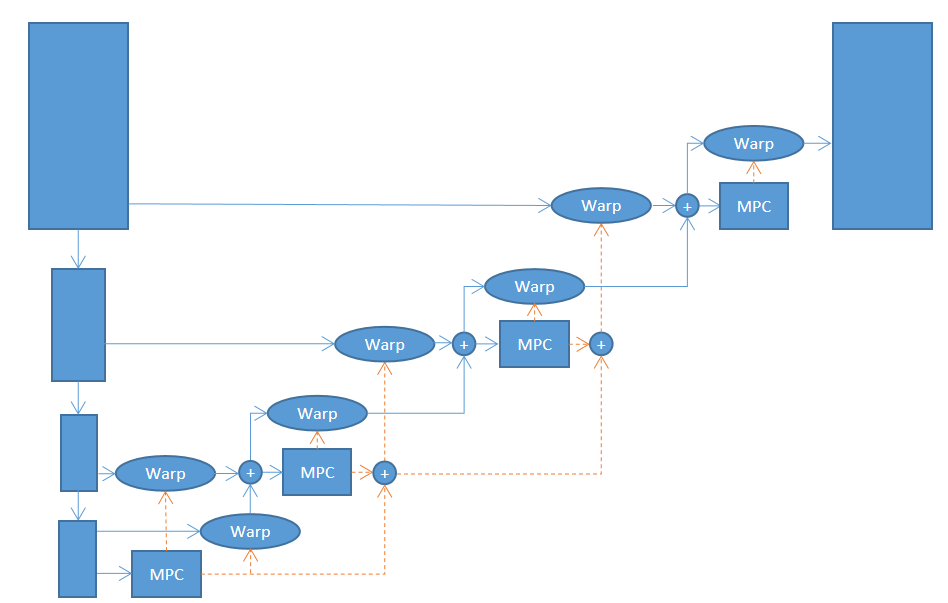}
    \\
    \figcspace
    \caption{
        \textbf{DeepBlueAI team (Track 2)}. Pyramid Correlation Alignment for Video Spatio-Temporal Super-Resolution
    }
	\label{fig:deepblueai_track2}
	\figspace
\end{figure}

DeepBlueAI team proposes the model ensemble strategy.
There are three models to be ensembled: PCA + upsample, PCA + EDVR~\cite{wang2019edvr}, NoFlow + EDVR.
PCA, which stands for pyramid correlation alignment, plays roles for the feature alignment and it is used for the frame interpolation.
Figure~\ref{fig:deepblueai_track2} describes the PCA module.
First, the multi-level features are extracted from an image using CNN layers.
Multi patch correlation (MPC) layer, which consists of patch correlation layer and convolution layer, calculates the offsets and the grid sampling layer uses the offset to warp the image, similar to spatial transformer network~\cite{jaderberg2015spatial}.
After the interpolation is done, they use EDVR network for the video super-resolution.
This method corresponds to PCA + EDVR model.
PCA + upsample model does not use EDVR, but they consider upsampled the output of PCA the final result.
NoFlow + EDVR substitutes PCA module to the simple convolution layers to get the inerpolated result.

\subsection{Team Horizon}

\begin{figure}[h]
	\centering
    \includegraphics[width=\linewidth]{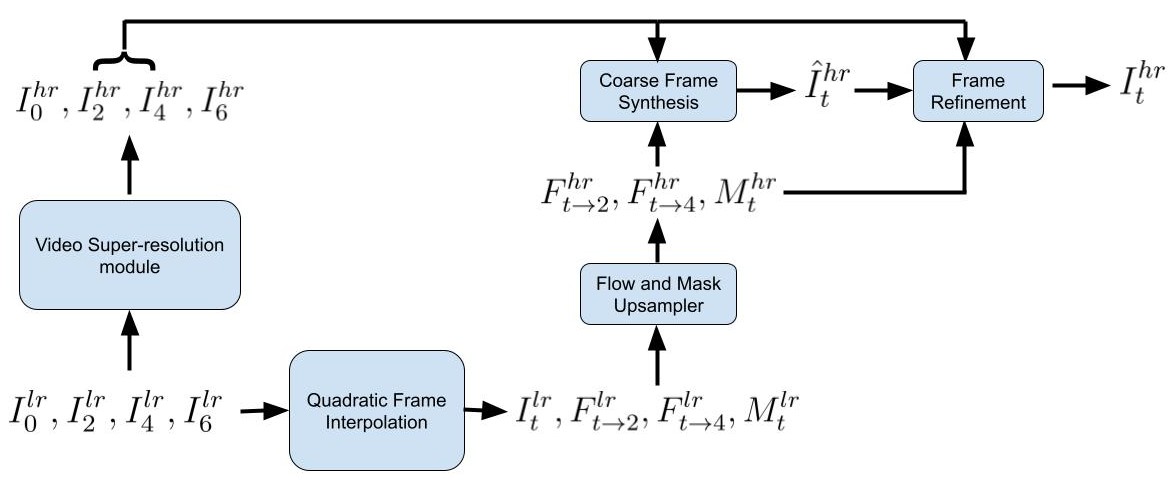}
    \\
    \figcspace
    \caption{
        \textbf{Team Horizon (Track 2)}. Efficient Space-time Super-Resolution using Flow Upsampling
    }
	\label{fig:horizon_track2}
	\figspace
\end{figure}

Team Horizon proposes efficient space-time super-resolution using flow upsampling method.
Given the 4 consecutive LR frames, they use quadratic frame interpolation~\cite{xu2019quadratic} model to get the center odd frame image as well as the flow maps and the blending mask.
Also, they adopt RSDN network~\cite{isobe2020video_cvpr} for video super-resolution.
Note that, they apply RSDN to the even frames only.
For estimating the HR-size odd frames, they use the output of quadratic video interpolation.
The upsampled flowmaps, upsampled blending mask and super-resolved even frames are used to make the HR images of odd frame indices.
The whole process is illustrated in Figure~\ref{fig:horizon_track2}.
More details could be found in \cite{dutta2021efficient}.

\subsection{MiGMaster\_XDU}

\begin{figure}[h]
	\centering
    \includegraphics[width=\linewidth]{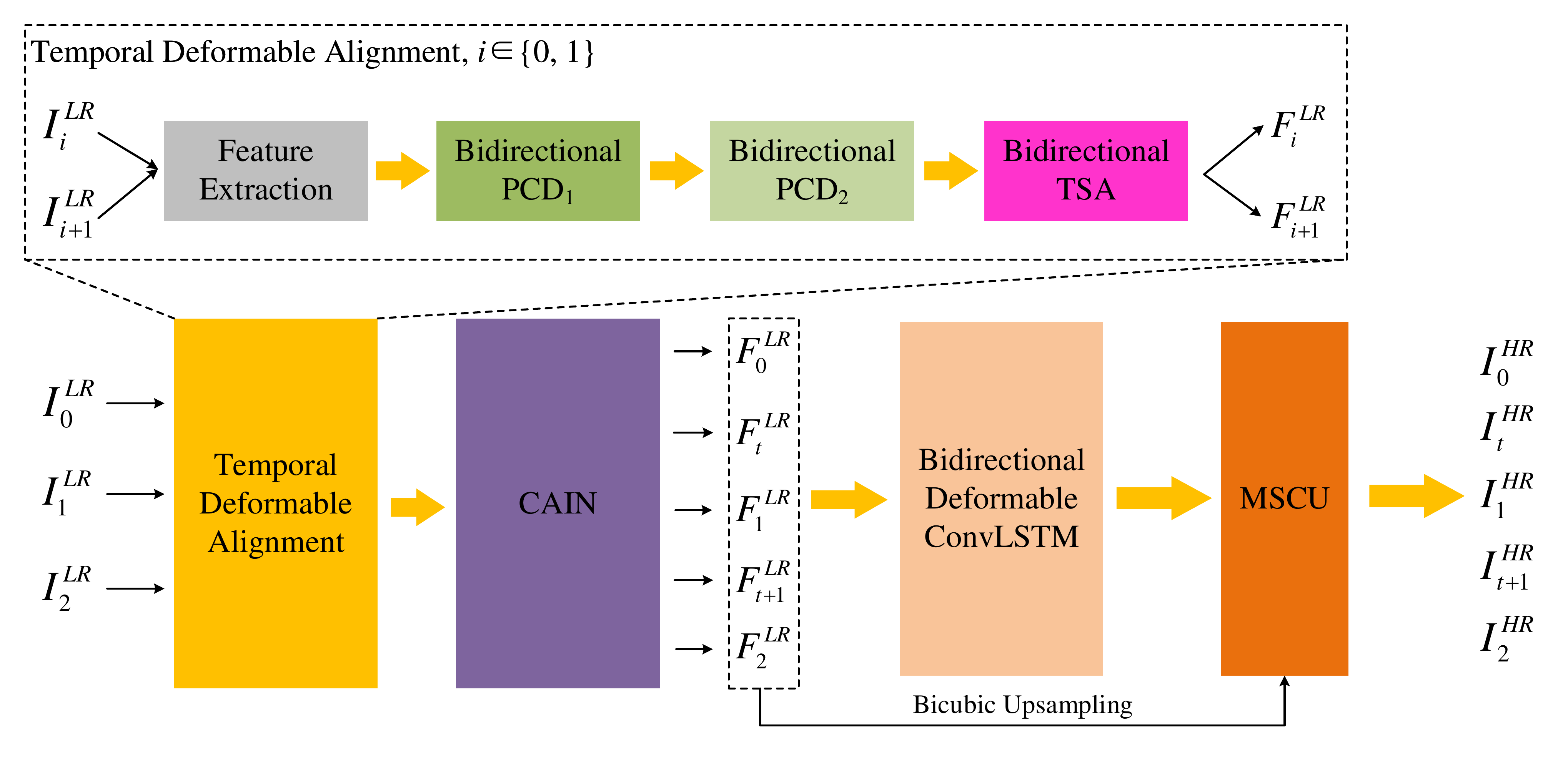}
    \\
    \figcspace
    \caption{
        \textbf{MiGMaster\_XDU team (Track 2)}. Multi-Stage Deformable Spatio-Temporal Video Super-Resolution
    }
	\label{fig:migmaster_track2}
	\figspace
\end{figure}

MiGMaster\_XDU team proposes multi-stage deformable spatio-temporal video super-resolution framework.
The main contribution comes from the temporal deformable alignment (TDA) module.
They expand the PCD, TSA modules of EDVR with the recurrent neural network.
By using bidirectional PCD, TSA modules, the features of neighbor frames can be aligned to any time steps, not only to the center frame index.
They use two bidirectional PCD modules to achieve coarse-to-fine temporal feature alignment.
After the alignment, the outputs of TDA module are sent to CAIN~\cite{choi2020channel} model, and they use bidirectional deformable convLSTM~\cite{xiang2020zooming} for feature aggregation.
Finally MSCU~\cite{liu2021large} network is adopted for video super-resoltuion.
The whole process is shown in Figure~\ref{fig:migmaster_track2}.

\begin{figure}[h]
	\centering
    \includegraphics[width=\linewidth]{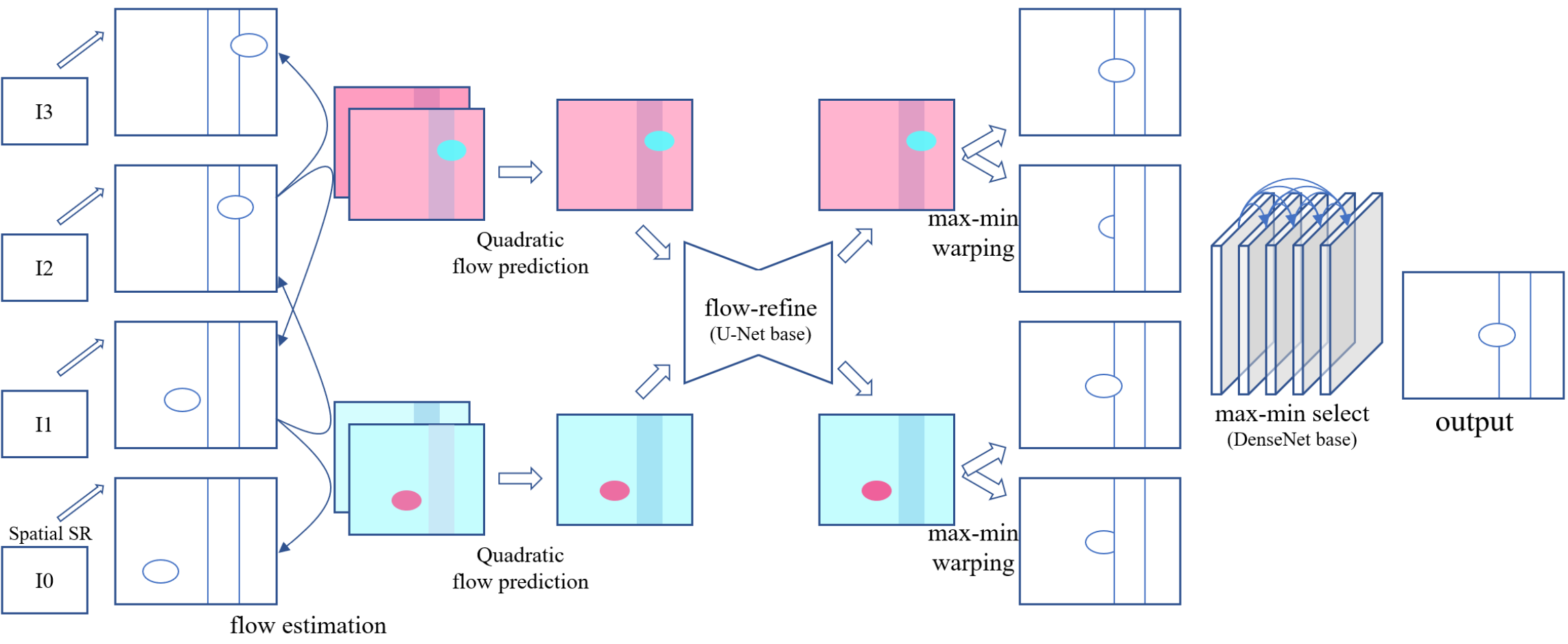}
    \\
    \figcspace
    \caption{
        \textbf{superbeam team (Track 2)}. Video Interpolation Using Deep Motion Selection Network / min-max select net
    }
	\label{fig:superbeam_track2}
	\figspace
\end{figure}

\subsection{superbeam}

superbeam team proposes a framework including flow refinement, max-min warping and max-min select.
The given frames are put into RCAN~\cite{zhang2018image} for super-resolution.
The flow maps are estimated using PWCNet~\cite{sun2018pwc} and refined by U-net architecture.
At warping stage, they apply max-min warping for overlapped regions.
The max warping means the value of the strongest motion is applied to the overlapped pixels, and the min warping means the value of the weakest motion is applied.
Finally the densenet structure, max-min selection is conducted and the final output is generated.
The procedure is illustrated in Figure~\ref{fig:superbeam_track2}

\subsection{DSST}

\begin{figure}[h]
	\centering
    \includegraphics[width=\linewidth]{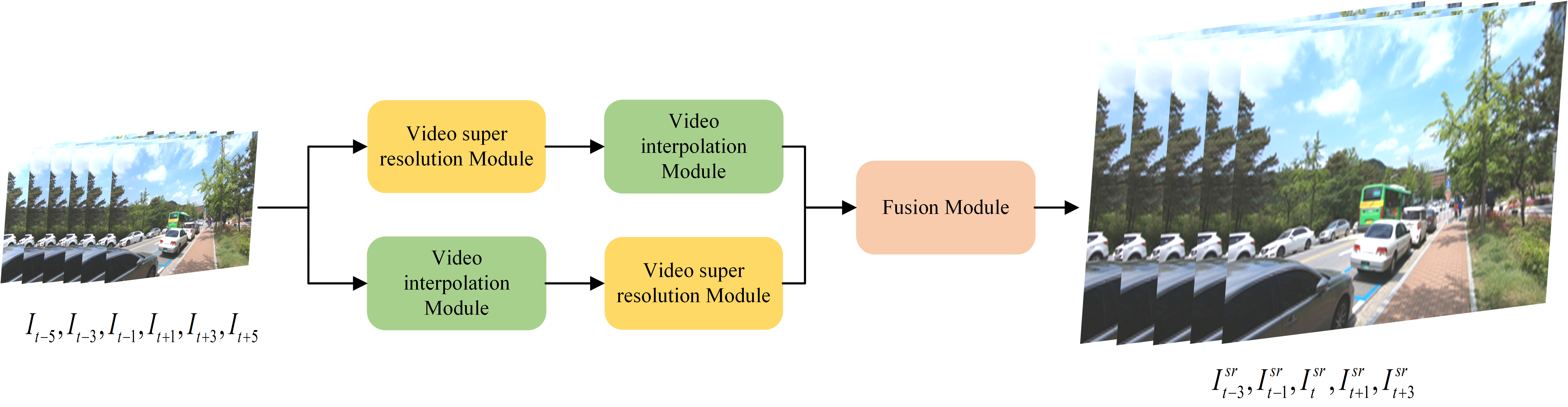}
    \\
    \figcspace
    \caption{
        \textbf{DSST team (Track 2)}. Dual-Stream Spatio-Temporal Video Enhancement Network
    }
	\label{fig:sjtu_track2}
	\figspace
\end{figure}

DSST team proposes the fusion network of two frameworks.
The both frameworks consist of video frame interpolation and video super-resolution, but one of them processes interpolation first and the other processes super-resolution first.
They adopt enhanced quadratic video interpolation model~\cite{liu2020enhanced} for the frame interpolation.
The architecture of the super-resolution network is similar to EDVR~\cite{wang2019edvr}, but the PCD module is replaced with flow align module.
The flow map outputs of the frame interpolation is refined by the flow refinement network and the images are aligned using the flow information.
The fusion module combines the two results to achieve higher performance.
Given the two images of two network streams, they generate the mask map and the final output is generated by the weighted sum of the two images.
The whole process is depicted in Figure~\ref{fig:sjtu_track2}.

\section*{Acknowledgments}

We thank the NTIRE 2021 sponsors: HUAWEI Technologies Co. Ltd., Wright Brothers Institute, Facebook Reality Labs, MediaTek Inc., OPPO Mobile Corp., Ltd. and ETH Zurich (Computer Vision Lab).

\appendix
\section{Teams and affiliations}
\label{sec:appendix}
\subsection*{NTIRE 2021 team}
\noindent\textit{\textbf{Title: }} NTIRE 2021 Challenge on Video Super-Resolution\\
\noindent\textit{\textbf{Members: }} \textit{Sanghyun Son$^1$ (thstkdgus35@snu.ac.kr)}, Suyoung Lee$^1$, Seungjun Nah$^1$, Radu Timofte$^2$,  Kyoung Mu Lee$^1$\\
\noindent\textit{\textbf{Affiliations: }}\\
$^1$ Department of ECE, ASRI, SNU, Korea\\
$^2$ Computer Vision Lab, ETH Zurich, Switzerland\\

\subsection*{NTU-SLab}
\noindent\textit{\textbf{Title: }}BasicVSR++\\
\noindent\textit{\textbf{Members: }}\textit{Kelvin C.K. Chan (chan0899@e.ntu.edu.sg)}, Shangchen Zhou, Xiangyu Xu, Chen Change Loy\\
\noindent\textit{\textbf{Affiliations: }}\\
S-Lab, Nanyang Technological University, Singapore\\

\subsection*{Imagination - Track 1}
\noindent\textit{\textbf{Title: }}Local to Context Video Super-Resolution\\
\noindent\textit{\textbf{Members: }}\textit{Chuming Lin$^1$ (chuminglin@tencent.com)}, Yuchun Dong$^2$, Boyuan Jiang$^1$, Donghao Luo$^1$, Siqian Yang$^1$, Ying Tai$^1$, Chengjie Wang$^1$, Jilin Li$^1$, Feiyue Huang$^1$\\
\noindent\textit{\textbf{Affiliations: }}\\
$^1$ YouTu Lab, Tencent\\
$^2$ East China Normal University, China\\

\subsection*{Imagination - Track 2}
\noindent\textit{\textbf{Title: }}Local to Context and Multi-level Quadratic model for video spatial and temporal super-resolution\\
\noindent\textit{\textbf{Members: }}\textit{Boyuan Jiang$^1$ (byronjiang@tencent.com)}, Chuming Lin$^1$, Xiaozhong Ji$^1$, Yuchun Dong$^2$, Donghao Luo$^1$, Wenqing Chu$^1$ Ying Tai$^1$, Chengjie Wang$^1$, Jilin Li$^1$, Feiyue Huang$^1$\\
\noindent\textit{\textbf{Affiliations: }}\\
$^1$ YouTu Lab, Tencent\\
$^2$ East China Normal University, China\\

\subsection*{model}
\noindent\textit{\textbf{Title: }}Flow-Alignment + Bi-directional Encoding + Adaptive Up-sampling\\
\noindent\textit{\textbf{Members: }}\textit{Chengpeng Chen$^1$ (chenchengpeng@megvii.com)}, Xiaojie Chu$^2$, Jie Zhang$^3$, Xin Lu$^1$, Liangyu Chen$^1$\\
\noindent\textit{\textbf{Affiliations: }}\\
$^1$ Megvii\\
$^2$ Peking University, China\\
$^3$ Fudan University, China\\

\subsection*{Noah Hisilicon SR}
\noindent\textit{\textbf{Title: }}Local and Global Feature Fusion Network\\
\noindent\textit{\textbf{Members: }}\textit{Xueyi Zou$^1$ (zouxueyi@huawei.com)}, Jing Lin$^1$, Guodong Du$^2$, Jia Hao$^2$\\
\noindent\textit{\textbf{Affiliations: }}\\
$^1$ Noah’s Ark Lab, Huawei\\
$^2$ HiSilicon(Shanghai) Technologies Co., Ltd.\\

\subsection*{VUE}
\noindent\textit{\textbf{Title: }}Two-Stage BasicVSR framework for Video Spatial Super-Resolution\\
\noindent\textit{\textbf{Members: }}\textit{Qi Zhang (zhangqi44@baidu.com)}, Lielin Jiang, Xin Li, He Zheng, Fanglong Liu, Dongliang He, Fu Li, Qingqing Dang\\
\noindent\textit{\textbf{Affiliations: }}\\
Department of Computer Vision Technology (VIS), Baidu Inc.\\

\subsection*{NERCMS}
\noindent\textit{\textbf{Title: }}Omniscient Video Super-Resolution\\
\noindent\textit{\textbf{Members: }}\textit{Peng Yi$^1$ (yipeng@whu.edu.cn)}, Zhongyuan Wang$^1$, Kui Jiang$^1$, Junjun Jiang$^2$, Jiayi Ma$^3$\\
\noindent\textit{\textbf{Affiliations: }}\\
$^1$ National Engineering Research Center for Multimedia Software, Wuhan University, China\\
$^2$ School of Computer Science and Technology, Harbin Institute of Technology, China\\
$^3$ Electronic Information School, Wuhan University, China\\

\subsection*{Diggers}
\noindent\textit{\textbf{Title: }}basicVSR\\
\noindent\textit{\textbf{Members: }}\textit{Yuxiang Chen (chenyx.cs@gmail.com)}, Yutong Wang\\
\noindent\textit{\textbf{Affiliations: }}\\
School of Computer Science and Engineering, University of Electronic Science and Technology of China, China\\


\subsection*{MT.Demacia}
\noindent\textit{\textbf{Title: }}AlignRefineNetwork\\
\noindent\textit{\textbf{Members: }}\textit{Ting Liu, (lt@meitu.com)}, Qichao Sun, Huanwei Liang\\
\noindent\textit{\textbf{Affiliations: }}\\
MTlab, Meitu, Inc.\\

\subsection*{MiG\_CLEAR}
\noindent\textit{\textbf{Title: }}An Improved EDVR model with SCNet and Group attention\\
\noindent\textit{\textbf{Members: }}\textit{Yiming Li (lym534701477@gmail.com)}, Zekun Li, Zhubo Ruan, Fanjie Shang, Chen Guo, Haining Li, Renjun Luo, Longjie Shen\\
\noindent\textit{\textbf{Affiliations: }}\\
Xidian university, China\\

\subsection*{VCL\_super\_resolution}
\noindent\textit{\textbf{Title: }}Quantized Warping and Residual Temporal Integration for Video Super-Resolution on Fast Motions\\
\noindent\textit{\textbf{Members: }}\textit{Kassiani Zafirouli (cassie.zaf@iti.gr)}, Konstantinos Karageorgos, Konstantinos Konstantoudakis, Anastasios Dimou, Petros Daras\\
\noindent\textit{\textbf{Affiliations: }}\\
Visual Computing Lab~(VCL), ITI/CERTH, Greece\\

\subsection*{SEU\_SR}
\noindent\textit{\textbf{Title: }}Recurrent Back-Projection Network\\
\noindent\textit{\textbf{Members: }}\textit{Xiaowei Song (220191705@seu.edu.cn)}, Xu Zhuo, Hanxi Liu\\
\noindent\textit{\textbf{Affiliations: }}\\
Southeast University, China\\

\subsection*{sVSRFI}
\noindent\textit{\textbf{Title: }}A two stage method for video spatio-temporal super-resolution\\
\noindent\textit{\textbf{Members: }}\textit{Yu Li$^1$ (ianyli@tencent.com)}, Ye Zhu$^2$\\
\noindent\textit{\textbf{Affiliations: }}\\
$^1$ Applied Research Center(ARC) Tencent PCG\\
$^2$ South China University of Technology, China\\

\subsection*{TheLastWaltz}
\noindent\textit{\textbf{Title: }}Separated framework for spatio temporal super resolution\\
\noindent\textit{\textbf{Members: }}\textit{Qing Wang$^1$ $^2$ (wangq529@mail2.sysu.edu.cn)}, Shijie Zhao$^1$, Xiaopeng Sun$^1$, Gen Zhan$^1$, Mengxi Guo$^1$, Junlin Li$^1$\\
\noindent\textit{\textbf{Affiliations: }}\\
$^1$ ByteDance\\
$^2$ Sun Yat-Sen University\\

\subsection*{T955}
\noindent\textit{\textbf{Title: }}BasicVSR after FLAVR\\
\noindent\textit{\textbf{Members: }}\textit{Tangxin Xie (xxh96@outlook.com)}, Yu Jia\\
\noindent\textit{\textbf{Affiliations: }}\\
China Electronic Technology Cyber Security Co., Ltd.\\

\subsection*{BOE-IOT-AIBD}
\noindent\textit{\textbf{Title: }}Two-Stages Video Spatial \& Temporal Super-Resolution Algorithm\\
\noindent\textit{\textbf{Members: }}\textit{Yunhua Lu (luyunhua@boe.com.cn)}, Wenhao Zhang, Mengdi Sun, Pablo Navarrete Michelini\\
\noindent\textit{\textbf{Affiliations: }}\\
AIoT R\&D Centre, BOE Technology Group Co., Ltd.\\

\subsection*{NaiveVSR}
\noindent\textit{\textbf{Title: }}Quadratic Space-Time Video Super-Resolution\\
\noindent\textit{\textbf{Members: }}\textit{Xueheng Zhang (zhangxueheng@sjtu.edu.cn)}, Hao Jiang, Zhiyu Chen, Li Chen\\
\noindent\textit{\textbf{Affiliations: }}\\
Shanghai Jiao Tong University, China\\

\subsection*{VIDAR}
\noindent\textit{\textbf{Title: }}Enhanced Temporal Alignment and Interpolation Network for Space-Time Video Super-Resolution\\
\noindent\textit{\textbf{Members: }}\textit{Zhiwei Xiong$^1$ (zwxiong@ustc.edu.cn)}, Zeyu Xiao$^1$, Ruikang Xu$^1$, Zhen Cheng$^1$, Xueyang Fu$^1$, Fenglong Song$^2$\\
\noindent\textit{\textbf{Affiliations: }}\\
$^1$ University of Science and Technology of China, China\\
$^2$ Huawei Noah’s Ark Lab\\

\subsection*{DeepBlueAI}
\noindent\textit{\textbf{Title: }}Pyramid Correlation Alignment For Video Spatio-Temporal Super-Resolution\\
\noindent\textit{\textbf{Members: }}\textit{Zhipeng Luo (luozp@deepblueai.com)}, Yuehan Yao\\
\noindent\textit{\textbf{Affiliations: }}\\
DeepBlue Technology Shanghai Co., Ltd.\\

\subsection*{Team Horizon}
\noindent\textit{\textbf{Title: }}Efficient Space-time Super Resolution using Flow Upsampling\\
\noindent\textit{\textbf{Members: }}\textit{Saikat Dutta$^1$ (saikat.dutta779@gmail.com)}, Nisarg A. Shah$^2$, Sourya Dipta Das$^3$\\
\noindent\textit{\textbf{Affiliations: }}\\
$^1$ Indian Institute of Technology Madras, India\\
$^2$ Indian Institute of Technology Jodhpur, India\\
$^3$ Jadavpur University, India\\

\subsection*{MiGMaster\_XDU}
\noindent\textit{\textbf{Title: }}Multi-Stage Deformable Spatio-Temporal Video Super-Resolution\\
\noindent\textit{\textbf{Members: }}\textit{Peng Zhao (iPrayer\_@outlook.com)}, Yukai Shi, Hongying Liu, Fanhua Shang, Yuanyuan Liu, Fei Chen, Fangxu Yu, Ruisheng Gao, Yixin Bai\\
\noindent\textit{\textbf{Affiliations: }}\\
Xidian University, China\\

\subsection*{superbeam}
\noindent\textit{\textbf{Title: }}Video Interpolation Using Deep Motion Selection Network / min-max select net\\
\noindent\textit{\textbf{Members: }}\textit{Jeonghwan Heo (hur881122@hanyang.ac.kr)}\\
\noindent\textit{\textbf{Affiliations: }}\\
Hanyang University, Korea\\

\subsection*{CNN}
\noindent\textit{\textbf{Title: }}Space-Time-Aware Multi-Resolution Video Enhancement\\
\noindent\textit{\textbf{Members: }}\textit{Shijie Yue$^1$ (1161126955@qq.com)}, Chenghua Li$^2$, Jinjing Li$^3$, Qian Zheng,$^3$ Ruipeng Gang$^4$, Ruixia Song$^1$\\
\noindent\textit{\textbf{Affiliations: }}\\
$^1$ NCUT, China\\
$^2$ CASIA, China\\
$^3$ CUC, China\\
$^4$ NRTA, China\\

\subsection*{Darambit}
\noindent\textit{\textbf{Title: }}Multi-frame Feature Combination Network for Video Super Resolution\\
\noindent\textit{\textbf{Members: }}\textit{Seungwoo Wee (slike0910@hanyang.ac.kr)}, Jechang Jeong\\
\noindent\textit{\textbf{Affiliations: }}\\
Hanyang University, Korea\\

\subsection*{DSST}
\noindent\textit{\textbf{Title: }}Dual-Stream Spatio-Temporal Video Enhancement Network\\
\noindent\textit{\textbf{Members: }}\textit{Chen Li (lcjurrivh@sjtu.edu.cn)}, Xinning Chai, Geyingjie Wen, Li Song\\
\noindent\textit{\textbf{Affiliations: }}\\
Shanghai Jiao Tong University, China\\

\newpage
{\small
\bibliographystyle{ieee_fullname}
\bibliography{egbib}

\begin{thebibliography}{10}\itemsep=-1pt

\bibitem{abuolaim2021ntire}
Abdullah Abuolaim, Radu Timofte, Michael~S Brown, et~al.
\newblock {NTIRE 2021} challenge for defocus deblurring using dual-pixel
  images: Methods and results.
\newblock In {\em CVPR Workshops}, 2021.

\bibitem{ancuti2021ntire}
Codruta~O Ancuti, Cosmin Ancuti, Florin-Alexandru Vasluianu, Radu Timofte,
  et~al.
\newblock {NTIRE 2021} nonhomogeneous dehazing challenge report.
\newblock In {\em CVPR Workshops}, 2021.

\bibitem{Bao_2019_CVPR}
Wenbo Bao, Wei-Sheng Lai, Chao Ma, Xiaoyun Zhang, Zhiyong Gao, and Ming-Hsuan
  Yang.
\newblock Depth-aware video frame interpolation.
\newblock In {\em CVPR}, 2019.

\bibitem{MEMC-Net}
Wenbo Bao, Wei-Sheng Lai, Xiaoyun Zhang, Zhiyong Gao, and Ming-Hsuan Yang.
\newblock {MEMC}-{N}et: Motion estimation and motion compensation driven neural
  network for video interpolation and enhancement.
\newblock {\em IEEE TPAMI}, 2019.

\bibitem{bhat2021ntire}
Goutam Bhat, Martin Danelljan, Radu Timofte, et~al.
\newblock {NTIRE 2021} challenge on burst super-resolution: Methods and
  results.
\newblock In {\em CVPR Workshops}, 2021.

\bibitem{Caballero_2017_CVPR}
Jose Caballero, Christian Ledig, Andrew Aitken, Alejandro Acosta, Johannes
  Totz, Zehan Wang, and Wenzhe Shi.
\newblock Real-time video super-resolution with spatio-temporal networks and
  motion compensation.
\newblock In {\em CVPR}, 2017.

\bibitem{chan2020basicvsr}
Kelvin~CK Chan, Xintao Wang, Ke Yu, Chao Dong, and Chen~Change Loy.
\newblock Basic{VSR}: The search for essential components in video
  super-resolution and beyond.
\newblock {\em arXiv preprint arXiv:2012.02181}, 2020.

\bibitem{Chi_2020_ECCV}
Zhixiang Chi, Rasoul~Mohammadi Nasiri, Zheng Liu, Juwei Lu, Jin Tang, and
  Konstantinos~N Plataniotis.
\newblock All at {O}nce: Temporally adaptive multi-frame interpolation with
  advanced motion modeling.
\newblock In {\em ECCV}, 2020.

\bibitem{choi2020channel}
Myungsub Choi, Heewon Kim, Bohyung Han, Ning Xu, and Kyoung~Mu Lee.
\newblock Channel attention is all you need for video frame interpolation.
\newblock In {\em AAAI}, 2020.

\bibitem{dai2017deformable}
Jifeng Dai, Haozhi Qi, Yuwen Xiong, Yi Li, Guodong Zhang, Han Hu, and Yichen
  Wei.
\newblock Deformable convolutional networks.
\newblock In {\em ICCV}, 2017.

\bibitem{dutta2021efficient}
Saikat Dutta, Nisarg~A. Shah, and Anurag Mittal.
\newblock Efficient space-time video super resolution using low-resolution flow
  and mask upsampling.
\newblock In {\em CVPR Workshops}, 2021.

\bibitem{elhelou2021ntire}
Majed El~Helou, Ruofan Zhou, Sabine S\"usstrunk, Radu Timofte, et~al.
\newblock {NTIRE 2021} depth guided image relighting challenge.
\newblock In {\em CVPR Workshops}, 2021.

\bibitem{gu2021ntire}
Jinjin Gu, Haoming Cai, Chao Dong, Jimmy~S. Ren, Yu Qiao, Shuhang Gu, Radu
  Timofte, et~al.
\newblock {NTIRE 2021} challenge on perceptual image quality assessment.
\newblock In {\em CVPR Workshops}, 2021.

\bibitem{haris2019recurrent}
Muhammad Haris, Gregory Shakhnarovich, and Norimichi Ukita.
\newblock Recurrent back-projection network for video super-resolution.
\newblock In {\em CVPR}, 2019.

\bibitem{haris2020space}
Muhammad Haris, Greg Shakhnarovich, and Norimichi Ukita.
\newblock Space-time-aware multi-resolution video enhancement.
\newblock In {\em CVPR}, 2020.

\bibitem{isobe2020video}
Takashi Isobe, Xu Jia, Shuhang Gu, Songjiang Li, Shengjin Wang, and Qi Tian.
\newblock Video super-resolution with recurrent structure-detail network.
\newblock In {\em ECCV}, 2020.

\bibitem{isobe2020video_cvpr}
Takashi Isobe, Songjiang Li, Xu Jia, Shanxin Yuan, Gregory Slabaugh, Chunjing
  Xu, Ya-Li Li, Shengjin Wang, and Qi Tian.
\newblock Video super-resolution with temporal group attention.
\newblock In {\em CVPR}, 2020.

\bibitem{jaderberg2015spatial}
Max Jaderberg, Karen Simonyan, Andrew Zisserman, and Koray Kavukcuoglu.
\newblock Spatial transformer networks.
\newblock In {\em NIPS}, 2015.

\bibitem{Jiang_2018_CVPR}
Huaizu Jiang, Deqing Sun, Varun Jampani, Ming-Hsuan Yang, Erik Learned-Miller,
  and Jan Kautz.
\newblock Super {S}lo{M}o: High quality estimation of multiple intermediate
  frames for video interpolation.
\newblock In {\em CVPR}, 2018.

\bibitem{jo2018deep}
Younghyun Jo, Seoung~Wug Oh, Jaeyeon Kang, and Seon~Joo Kim.
\newblock Deep video super-resolution network using dynamic upsampling filters
  without explicit motion compensation.
\newblock In {\em CVPR}, 2018.

\bibitem{kalluri2020flavr}
Tarun Kalluri, Deepak Pathak, Manmohan Chandraker, and Du Tran.
\newblock {FLAVR}: Flow-agnostic video representations for fast frame
  interpolation.
\newblock {\em arXiv preprint arXiv:2012.08512}, 2020.

\bibitem{Lee_2020_CVPR}
Hyeongmin Lee, Taeoh Kim, Tae-young Chung, Daehyun Pak, Yuseok Ban, and
  Sangyoun Lee.
\newblock Ada{C}of: Adaptive collaboration of flows for video frame
  interpolation.
\newblock In {\em CVPR}, 2020.

\bibitem{qvi_iccvw19}
Siyao Li, Xiangyu Xu, Ze Pan, and Wenxiu Sun.
\newblock Quadratic video interpolation for {VTSR} challenge.
\newblock In {\em ICCV Workshops}, 2019.

\bibitem{li2015space}
Tao Li, Xiaohai He, Qizhi Teng, Zhengyong Wang, and Chao Ren.
\newblock Space--time super-resolution with patch group cuts prior.
\newblock {\em Signal Processing: Image Communication}, 30:147--165, 2015.

\bibitem{li2020mucan}
Wenbo Li, Xin Tao, Taian Guo, Lu Qi, Jiangbo Lu, and Jiaya Jia.
\newblock Mu{C}an: Multi-correspondence aggregation network for video
  super-resolution.
\newblock In {\em ECCV}, 2020.

\bibitem{Lim_2017_CVPR_Workshops}
Bee Lim, Sanghyun Son, Heewon Kim, Seungjun Nah, and Kyoung~Mu Lee.
\newblock Enhanced deep residual networks for single image super-resolution.
\newblock In {\em CVPR Workshops}, 2017.

\bibitem{liu2021large}
Hongying Liu, Peng Zhao, Zhubo Ruan, Fanhua Shang, and Yuanyuan Liu.
\newblock Large motion video super-resolution with dual subnet and multi-stage
  communicated upsampling.
\newblock {\em arXiv preprint arXiv:2103.11744}, 2021.

\bibitem{liu2021ntire}
Jerrick Liu, Oliver Nina, Radu Timofte, et~al.
\newblock {NTIRE 2021} multi-modal aerial view object classification challenge.
\newblock In {\em CVPR Workshops}, 2021.

\bibitem{Liu_2020_CVPR_Workshops}
Shuai Liu, Chenghua Li, Nan Nan, Ziyao Zong, and Ruixia Song.
\newblock {MMDM}: Multi-frame and multi-scale for image demoireing.
\newblock In {\em CVPR Workshops}, June 2020.

\bibitem{liu2020enhanced}
Yihao Liu, Liangbin Xie, Li Siyao, Wenxiu Sun, Yu Qiao, and Chao Dong.
\newblock Enhanced quadratic video interpolation.
\newblock In {\em ECCV Workshops}, 2020.

\bibitem{Liu_2017_ICCV}
Ziwei Liu, Raymond~A. Yeh, Xiaoou Tang, Yiming Liu, and Aseem Agarwala.
\newblock Video frame synthesis using deep voxel flow.
\newblock In {\em ICCV}, 2017.

\bibitem{long2016learning}
Gucan Long, Laurent Kneip, Jose~M Alvarez, Hongdong Li, Xiaohu Zhang, and
  Qifeng Yu.
\newblock Learning image matching by simply watching video.
\newblock In {\em ECCV}, 2016.

\bibitem{lugmayr2021ntire}
Andreas Lugmayr, Martin Danelljan, Radu Timofte, et~al.
\newblock {NTIRE 2021} learning the super-resolution space challenge.
\newblock In {\em CVPR Workshops}, 2021.

\bibitem{Meyer_2018_CVPR}
Simone Meyer, Abdelaziz Djelouah, Brian McWilliams, Alexander Sorkine-Hornung,
  Markus Gross, and Christopher Schroers.
\newblock Phase{N}et for video frame interpolation.
\newblock In {\em CVPR}, 2018.

\bibitem{Meyer_2015_CVPR}
Simone Meyer, Oliver Wang, Henning Zimmer, Max Grosse, and Alexander
  Sorkine-Hornung.
\newblock Phase-{B}ased frame interpolation for video.
\newblock In {\em CVPR}, 2015.

\bibitem{mudenagudi2010space}
Uma Mudenagudi, Subhashis Banerjee, and Prem~Kumar Kalra.
\newblock Space-time super-resolution using graph-cut optimization.
\newblock {\em IEEE TPAMI}, 33(5):995--1008, 2010.

\bibitem{Nah_2019_CVPR_Workshops_REDS}
Seungjun Nah, Sungyong Baik, Seokil Hong, Gyeongsik Moon, Sanghyun Son, Radu
  Timofte, and Kyoung~Mu Lee.
\newblock {NTIRE} 2019 challenges on video deblurring and super-resolution:
  Dataset and study.
\newblock In {\em CVPR Workshops}, 2019.

\bibitem{Nah_2020_ECCV_Workshops_VTSR}
Seungjun Nah, Sanghyun Son, Jaerin Lee, Radu Timofte, Kyoung~Mu Lee, et~al.
\newblock {AIM} 2020 challenge on video temporal super-resolution.
\newblock In {\em ECCV Workshops}, 2020.

\bibitem{nah2021ntire}
Seungjun Nah, Sanghyun Son, Suyoung Lee, Radu Timofte, Kyoung~Mu Lee, et~al.
\newblock {NTIRE 2021} challenge on image deblurring.
\newblock In {\em CVPR Workshops}, 2021.

\bibitem{Nah_2020_CVPR_Workshops_Deblur}
Seungjun Nah, Sanghyun Son, Radu Timofte, and Kyoung~Mu Lee.
\newblock {NTIRE} 2020 challenge on image and video deblurring.
\newblock In {\em CVPR Workshops}, 2020.

\bibitem{Nah_2019_ICCV_Workshops_VTSR}
Seungjun Nah, Sanghyun Son, Radu Timofte, Kyoung~Mu Lee, et~al.
\newblock {AIM} 2019 challenge on video temporal super-resolution: Methods and
  results.
\newblock In {\em ICCV Workshops}, 2019.

\bibitem{Nah_2019_CVPR_Workshops_Deblur}
Seungjun Nah, Radu Timofte, Sungyong Baik, Seokil Hong, Gyeongsik Moon,
  Sanghyun Son, and Kyoung~Mu Lee.
\newblock {NTIRE} 2019 challenge on video deblurring: Methods and results.
\newblock In {\em CVPR Workshops}, 2019.

\bibitem{Nah_2019_CVPR_Workshops_SR}
Seungjun Nah, Radu Timofte, Shuhang Gu, Sungyong Baik, Seokil Hong, Gyeongsik
  Moon, Sanghyun Son, and Kyoung~Mu Lee.
\newblock {NTIRE} 2019 challenge on video super-resolution: Methods and
  results.
\newblock In {\em CVPR Workshops}, 2019.

\bibitem{Niklaus_2018_CVPR}
Simon Niklaus and Feng Liu.
\newblock Context-aware synthesis for video frame interpolation.
\newblock In {\em CVPR}, 2018.

\bibitem{Niklaus_2020_CVPR}
Simon Niklaus and Feng Liu.
\newblock Softmax splatting for video frame interpolation.
\newblock In {\em CVPR}, 2020.

\bibitem{Niklaus_2017_CVPR}
Simon Niklaus, Long Mai, and Feng Liu.
\newblock Video frame interpolation via adaptive convolution.
\newblock In {\em CVPR}, 2017.

\bibitem{Park_2020_ECCV}
Junheum Park, Keunsoo Ko, Chul Lee, and Chang-Su Kim.
\newblock {BMBC}: Bilateral motion estimation with bilateral cost volume for
  video interpolation.
\newblock In {\em ECCV}, 2020.

\bibitem{Peleg_2019_CVPR}
Tomer Peleg, Pablo Szekely, Doron Sabo, and Omry Sendik.
\newblock Im-net for high resolution video frame interpolation.
\newblock In {\em CVPR}, 2019.

\bibitem{perez2021ntire}
Eduardo P\'erez-Pellitero, Sibi Catley-Chandar, Ale\v{s} Leonardis, Radu
  Timofte, et~al.
\newblock {NTIRE 2021} challenge on high dynamic range imaging: Dataset,
  methods and results.
\newblock In {\em CVPR Workshops}, 2021.

\bibitem{ranjan2017optical}
Anurag Ranjan and Michael~J Black.
\newblock Optical flow estimation using a spatial pyramid network.
\newblock In {\em CVPR}, 2017.

\bibitem{Reda_2019_ICCV}
Fitsum~A. Reda, Deqing Sun, Aysegul Dundar, Mohammad Shoeybi, Guilin Liu,
  Kevin~J. Shih, Andrew Tao, Jan Kautz, and Bryan Catanzaro.
\newblock Unsupervised video interpolation using cycle consistency.
\newblock In {\em ICCV}, 2019.

\bibitem{shechtman2002increasing}
Eli Shechtman, Yaron Caspi, and Michal Irani.
\newblock Increasing space-time resolution in video.
\newblock In {\em ECCV}, 2002.

\bibitem{shechtman2005space}
Eli Shechtman, Yaron Caspi, and Michal Irani.
\newblock Space-time super-resolution.
\newblock {\em IEEE TPAMI}, 27(4):531--545, 2005.

\bibitem{Son_2020_ECCV_Workshops}
Sanghyun Son, Jaerin Lee, Seungjun Nah, Radu Timofte, and Kyoung~Mu Lee.
\newblock {AIM} 2020 challenge on video temporal super-resolution.
\newblock In {\em ECCV}, pages 23--40. Springer, 2020.

\bibitem{sun2018pwc}
Deqing Sun, Xiaodong Yang, Ming-Yu Liu, and Jan Kautz.
\newblock {PWC}-{N}et: {CNN}s for optical flow using pyramid, warping, and cost
  volume.
\newblock In {\em CVPR}, 2018.

\bibitem{Tao_2017_ICCV}
Xin Tao, Hongyun Gao, Renjie Liao, Jue Wang, and Jiaya Jia.
\newblock Detail-revealing deep video super-resolution.
\newblock In {\em ICCV}, 2017.

\bibitem{tian2018tdan}
Yapeng Tian, Yulun Zhang, Yun Fu, and Chenliang Xu.
\newblock {TDAN:} {T}emporally deformable alignment network for video
  super-resolution.
\newblock In {\em CVPR}, 2020.

\bibitem{wang2019edvr}
Xintao Wang, Kelvin~C.K. Chan, Ke Yu, Chao Dong, and Chen~Change Loy.
\newblock {EDVR}: Video restoration with enhanced deformable convolutional
  networks.
\newblock In {\em CVPR Workshops}, 2019.

\bibitem{wang2004image}
Zhou Wang, Alan~C Bovik, Hamid~R Sheikh, Eero~P Simoncelli, et~al.
\newblock Image quality assessment: from error visibility to structural
  similarity.
\newblock {\em TIP}, 13(4):600--612, 2004.

\bibitem{xiang2020zooming}
Xiaoyu Xiang, Yapeng Tian, Yulun Zhang, Yun Fu, Jan~P Allebach, and Chenliang
  Xu.
\newblock Zooming slow-mo: Fast and accurate one-stage space-time video
  super-resolution.
\newblock In {\em CVPR}, 2020.

\bibitem{xiao2020space}
Zeyu Xiao, Zhiwei Xiong, Xueyang Fu, Dong Liu, and Zheng-Jun Zha.
\newblock Space-time video super-resolution using temporal profiles.
\newblock In {\em ACM Multimedia}, pages 664--672, 2020.

\bibitem{xu2019quadratic}
Xiangyu Xu, Li Siyao, Wenxiu Sun, Qian Yin, and Ming~Hsuan Yang.
\newblock Quadratic video interpolation.
\newblock {\em NeurIPS}, 2019.

\bibitem{xue2019video}
Tianfan Xue, Baian Chen, Jiajun Wu, Donglai Wei, and William~T Freeman.
\newblock Video enhancement with task-oriented flow.
\newblock {\em IJCV}, 127(8):1106--1125, 2019.

\bibitem{yang2021ntire}
Ren Yang, Radu Timofte, et~al.
\newblock {NTIRE 2021} challenge on quality enhancement of compressed video:
  Methods and results.
\newblock In {\em CVPR Workshops}, 2021.

\bibitem{Yuan_2019_CVPR}
Liangzhe Yuan, Yibo Chen, Hantian Liu, Tao Kong, and Jianbo Shi.
\newblock Zoom-{I}n-{T}o-{C}heck: Boosting video interpolation via
  instance-level discrimination.
\newblock In {\em CVPR}, 2019.

\bibitem{zhang2018unreasonable}
Richard Zhang, Phillip Isola, Alexei~A Efros, Eli Shechtman, and Oliver Wang.
\newblock The unreasonable effectiveness of deep features as a perceptual
  metric.
\newblock In {\em CVPR}, 2018.

\bibitem{zhang2018image}
Yulun Zhang, Kunpeng Li, Kai Li, Lichen Wang, Bineng Zhong, and Yun Fu.
\newblock Image super-resolution using very deep residual channel attention
  networks.
\newblock In {\em ECCV}, 2018.

\bibitem{zuckerman2020across}
Liad~Pollak Zuckerman, Eyal Naor, George Pisha, Shai Bagon, and Michal Irani.
\newblock Across scales and across dimensions: Temporal super-resolution using
  deep internal learning.
\newblock In {\em ECCV}, 2020.

\end{thebibliography}
}

\end{document}